\documentclass[journal]{IEEEtran}

\usepackage{xcolor,soul,framed} 

\colorlet{shadecolor}{yellow}
\usepackage[pdftex]{graphicx}
\graphicspath{{../pdf/}{../jpeg/}}
\DeclareGraphicsExtensions{.pdf,.jpeg,.png}

\usepackage[cmex10]{amsmath}
\usepackage{array}
\usepackage{mdwmath}
\usepackage{mdwtab}
\usepackage{eqparbox}
\usepackage{url}
\usepackage{multirow}
\usepackage{amssymb}             
\usepackage{amsfonts}            
\usepackage{mathrsfs}            
\usepackage{booktabs}
\usepackage{tabularx}
\usepackage{adjustbox}
\usepackage[switch]{lineno}
\usepackage{algorithm}
\usepackage{algpseudocode}
\usepackage{placeins}
\algrenewcommand\algorithmicrequire{\textbf{Input:}}
\algrenewcommand\algorithmicensure{\textbf{Output:}}
\usepackage{makecell}
\usepackage{graphicx}
\usepackage{hyperref}
\hypersetup{
colorlinks=true,
linkcolor=blue,
citecolor=blue,
urlcolor=black
}
\usepackage[numbers,sort&compress]{natbib}
\begin{document}
\bstctlcite{IEEEexample:BSTcontrol}
    \title{SeaCausal-FL: Federated Fuzzy Causal Learning for Maritime IoT Fault Diagnosis and Counterfactual Reasoning}
  \author{Yuhang Qiu, Haihan Zhu, Koteeswaran Seerangan, Longsheng Zhu, Xiong Wang, Yijun Lu, Zheng Lin, \\ Fangmin Ren and Jialiang Xie
  \thanks{Yuhang Qiu and Haihan Zhu contributed equally to this work. This work was supported in part by the National Natural Science Foundation of China (No. 12571585) and Scientific Research Start-up Program of Jimei University (No. Z8268125).}
  {\thanks{
  Yuhang Qiu, Fangmin Ren, and Jialiang Xie are with the School of Science, Jimei University, Xiamen 361021, China. (e-mail: yuhang.qiu@jmu.edu.cn; fangminren@jmu.edu.cn; xiejialiang@jmu.edu.cn).}

   \thanks{Haihan Zhu is with the School of Artificial Intelligence, Hebei University of Technology, Tianjin 300401, China (email: 242794@stu.hebut.edu.cn)}

  \thanks{Koteeswaran Seerangan is with the Department of Computer Science and Engineering,  R. M. K. Engineering College (Autonomous), Kavaraipettai - 601206, Tamil Nadu, India (e-mail: skn.cse@rmkec.ac.in).}

  \thanks{Longsheng Zhu is with the College of Information Network Security, People’s Public Security University of China, Beijing 100038, China (e-mail: 2024111028@stu.ppsuc.edu.cn).}

\thanks{Xiong Wang is with the School of Computer Science and Technology, University of Science and Technology of China,
Hefei 230026, China (e-mail: wangxiong@ustc.edu.cn).}

\thanks{Yijun Lu is with the Department of Computer Science and Engineering, Waseda University, Tokyo,  Japan (e-mail:
yijun@ruri.waseda.jp).}

 \thanks{Zheng Lin is with the Interdisciplinary Centre for Security, Reliability and Trust (SnT), University of Luxembourg, Luxembourg (e-mail: zhenglin@ieee.org).}

  \thanks{(Corresponding author: Jialiang Xie)}

}

}

\maketitle

\begin{abstract} 
Reliable marine-engine fault diagnosis in maritime IoT is challenged by distributed data ownership, heterogeneous fault distributions, and continuously changing operating conditions. This paper proposes SeaCausal-FL, a federated fuzzy causal learning framework that combines a shared temporal diagnostic path with mechanism-conditioned causal reasoning. An interval type-2 fuzzy layer represents uncertain and overlapping operating mechanisms, while each mechanism is associated with a physics-constrained structural causal model. Before aggregation, locally learned mechanisms are aligned using operating context, causal structure, and conditional intervention--response signatures. Model parameters are then aggregated according to sample, class, mechanism, and mechanism--class evidence instead of client sample size alone. The learned structural equations further support interval counterfactual reasoning through abduction, action, and prediction. Experiments on a marine-engine fault dataset and a real-data-calibrated semi-synthetic causal benchmark show that SeaCausal-FL achieves an average F1-score of 87.07\% across four client partitions, with AUROC and AUPRC of 98.98\% and 94.81\%, respectively. It also maintains strong performance under unseen loads and fault-type omission during training. On the causal benchmark, SeaCausal-FL reaches an Edge-F1 of approximately 0.58 and an Edge-AUPRC of 0.68, reduces coefficient RMSE to about 0.14, and provides favorable counterfactual estimation and intervention decisions.

\end{abstract}


%
\IEEEpeerreviewmaketitle


\newpage
\section{Introduction}
\label{sec:introduction}

\IEEEPARstart{M}{arine} transportation is increasingly dependent on onboard sensing, edge computing, and shore-side intelligent services. Under the Internet of Ships and maritime Internet of Things paradigms, modern vessels routinely collect multivariate signals related to engine load, speed, fuel supply, cooling, combustion, exhaust conditions, and power output \cite{aslam2020internet,chen2025edge,huo2020cellular}. These measurements provide an important basis for marine-engine fault diagnosis by supporting the early detection of abnormal operating states before they develop into propulsion failures, which is important for navigation safety, maintenance cost reduction, and vessel availability \cite{youssef2024survey,kowalski2017fault,wang2021monitoring}.

Machine learning, as a powerful data-driven paradigm for extracting discriminative representations \cite{qiu2025ifvit,lin2024fedsn,fang2024ic3m,sun2025rrto,zhang2024satfed} has been widely applied to maritime component monitoring and marine diesel-engine diagnosis \cite{kowalski2017fault,wang2021monitoring,shi2025universal,zhao2025crossdomain}. However, diagnostic performance remains sensitive to operating-condition variation \cite{shi2025universal,zhao2025crossdomain,gao2024disentanglement}. The same fault can produce different sensor responses under different loads or thermal states, while different faults can cause similar changes in pressure, temperature, or flow. Moreover, operating conditions commonly vary continuously rather than forming clearly separated regimes. These characteristics create heterogeneous and overlapping fault patterns that are difficult to represent with a single condition-independent diagnostic model.

A further challenge arises from the distributed ownership of maritime data. Measurements collected from different vessels, engine units, maintenance organizations, or test campaigns are generally stored at separate sites, while direct data sharing can be restricted by commercial confidentiality, communication cost, and data-governance requirements. Federated learning provides a practical way to train diagnostic models without transferring raw measurements \cite{qian2025review, lin2025adaptsfl,chen2026irsaided}.  Existing federated fault-diagnosis studies have addressed data heterogeneity and domain shifts through transfer learning, domain generalization, meta-learning, and personalized aggregation \cite{zhang2021federated,zhang2022privacy,zhao2024fdg,han2024class,cui2026cooperative,xu2026openset,lin2026hasfl}. Hierarchical split federated learning has also been studied for coordinating model splitting and aggregation across multi-tier edge systems, further illustrating the importance of adapting distributed learning to heterogeneous computing environments \cite{lin2025hierarchical}. These methods can improve diagnostic robustness under several forms of data heterogeneity, but their coordination units are generally complete models, clients, domains, latent features, or fault-class representations.

Such coordination does not explicitly distinguish multiple uncertain operating mechanisms that can coexist within the same client. It also does not guarantee that locally learned components with the same parameter index represent physically corresponding operating mechanisms across clients. As a result, direct aggregation can mix operating-specific knowledge with different physical meanings. Class-aware aggregation alleviates label skew but does not determine whether the available class evidence is supported under the same operating mechanism. In addition, most federated diagnostic models are optimized primarily for fault recognition and do not preserve mechanism-specific physical relations required for intervention and counterfactual analysis. These limitations motivate a finer-grained framework that coordinates operating mechanisms rather than only complete client models or class-level representations.

To address these limitations, this paper proposes SeaCausal-FL, a federated fuzzy causal learning framework for maritime IoT fault diagnosis and counterfactual reasoning. SeaCausal-FL combines a shared temporal diagnostic path with an operating-dependent fuzzy causal path. An interval type-2 fuzzy layer \cite{liang2000interval,qiao2024interval} represents uncertain and overlapping operating mechanisms through lower, midpoint, and upper membership weights. Each mechanism parameterizes a structural causal model within a physics-constrained candidate graph \cite{pearl2009causal,scholkopf2021toward}, while its mechanism-specific output acts as a residual correction to the shared diagnosis. The residual contribution is initialized at zero so that the common diagnostic representation is established before mechanism-specific causal corrections become active.

Before aggregation, locally learned mechanisms are aligned with the broadcast global reference using operating context and causal structure. When the preliminary alignment indicates a possible permutation, intervention--response signatures are further used to refine the correspondence. Different parameter groups are then aggregated according to their supporting evidence: shared parameters use sample counts, class-specific parameters use class counts, and mechanism-related parameters use fuzzy membership and mechanism--class evidence. The learned structural equations further support counterfactual inference through abduction, action, and prediction, providing nominal and interval-valued fault-risk estimates under feasible physical interventions.

The main contributions of this paper are summarized as follows.
\begin{enumerate}
\item We propose a federated diagnostic architecture for simultaneous operating-condition and label heterogeneity. It combines a shared temporal diagnostic model with mechanism-conditioned causal residuals to preserve stable fault recognition while modeling operating-dependent physical responses.

\item We introduce an interval type-2 fuzzy causal mechanism model for uncertain and overlapping operating conditions. Each fuzzy mechanism parameterizes a structural causal model within a physics-constrained candidate graph, while its center, uncertainty footprint, edge evidence, and structural coefficients are learned from training data rather than manually defined operating thresholds or fixed causal weights.

\item We develop a physical-response-based mechanism alignment and mechanism class-aware aggregation strategy. Local mechanisms are matched according to operating context, causal structure, and intervention response, and their parameters are aggregated using effective fuzzy membership mass and class evidence rather than client sample size alone.

\item We evaluate SeaCausal-FL on a real marine-engine fault dataset and a real-data-calibrated semi-synthetic causal benchmark. The experiments cover natural non-IID, IID, and Dirichlet partitions, unseen operating loads, incomplete fault-type coverage, client participation, causal structure recovery, and counterfactual estimation.
\end{enumerate}

The remainder of this paper is organized as follows. Section II reviews related work. Section III presents the proposed SeaCausal-FL framework. Section IV describes the experimental settings and reports the results. Section V concludes the paper.


\begin{figure*}[!t]
\centering
\includegraphics[width=0.85\textwidth]{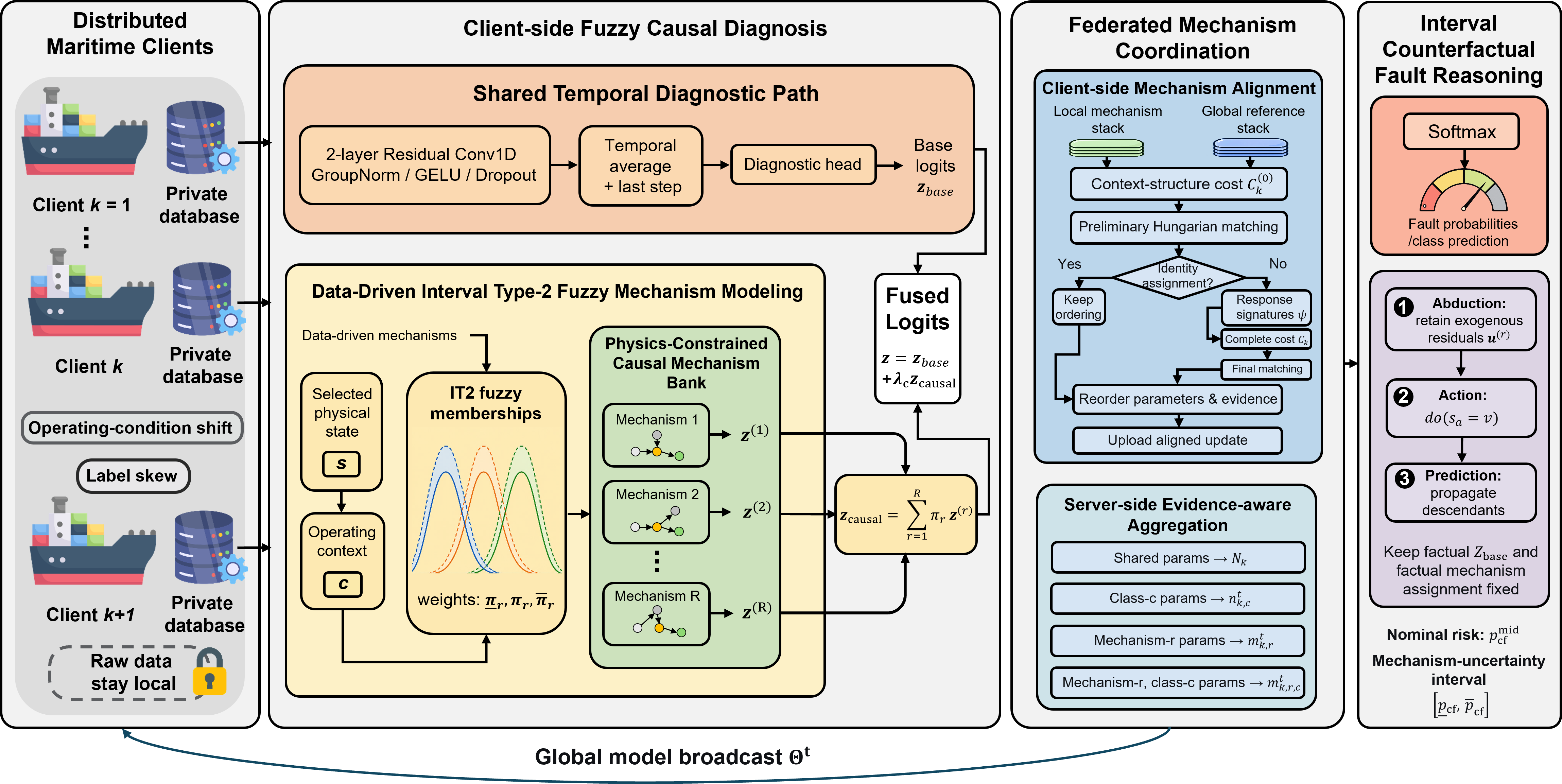}
\caption{Overall framework of SeaCausal-FL. Each maritime client performs client-side fuzzy causal diagnosis through a shared temporal diagnostic path and an operating-dependent fuzzy causal path. Locally updated mechanisms are aligned with the broadcast global reference before upload. The server then performs evidence-aware aggregation using sample, class, mechanism, and mechanism--class evidence. The resulting global model supports both fault diagnosis and interval counterfactual reasoning.}
\label{fig:framework}
\end{figure*}
\section{Related Work}
Marine engine fault diagnosis has gradually shifted from expert rules and conventional signal analysis to machine learning and deep learning methods \cite{han2021lstmvae,naryanto2023diesel,wang2023dpgcn,zhang2026dualpath}. Existing studies have used ensemble learning, manifold learning, recurrent networks, and transfer learning to extract diagnostic information from pressure, temperature, vibration, and other engine measurements \cite{kowalski2017fault,wang2021monitoring,shi2025universal}. Recent cross-condition methods further employ domain adaptation, feature disentanglement, and centroid alignment to reduce distribution differences among loads, machines, and operating environments \cite{zhao2025crossdomain,gao2024disentanglement}. These methods improve recognition under changing conditions, but they generally regard each load or machine as a predefined domain or seek representations that remove operating-condition information. Such treatment is less suitable when operating regimes change gradually, overlap with one another, or produce genuinely different physical responses to the same fault. Most of these methods also assume centralized access to the diagnostic data.

Federated fault diagnosis has received increasing attention. Early studies combined self-supervised local training with validation-guided aggregation, while federated transfer-learning methods used prior distributions, feature adaptation, and target self-adaptation to handle cross-machine and cross-condition shifts \cite{zhang2021federated,zhang2022privacy,li2023target}. Federated domain generalization and meta-learning have also been studied for unseen domains, limited local samples, and few-shot diagnosis \cite{zhao2024fdg,chen2023industrial,cui2023representation}. More recent methods use class information, personalized classifiers, and fault prototypes to address label-distribution skew \cite{han2024class,fan2025prototype}. Personalized aggregation has further been considered when a client contains several working conditions, while communication-aware and open-set methods have been developed for offshore equipment and unknown faults \cite{cui2026cooperative,lu2024event,xu2026openset}. These studies cover several important forms of data heterogeneity, but their coordination units remain complete models, clients, domains, latent features, or fault-class prototypes. They do not explicitly determine whether local components learned at different clients describe the same operating mechanism. Class-aware aggregation identifies available fault evidence, but it does not determine whether that evidence is obtained under a corresponding physical mechanism. As a result, operating-specific knowledge with different meanings can still be mixed during aggregation.

Fuzzy systems provide a natural way to describe operating regimes without imposing hard boundaries. In particular, interval type-2 fuzzy models represent uncertainty in the membership itself and have been applied to nonstationary industrial processes \cite{liang2000interval,qiao2024interval}. In existing diagnostic models, however, fuzzy memberships are mainly used for rule activation, sample weighting, feature mapping, or prediction confidence. They do not usually define distinct physical mechanisms. Structural causal models offer a complementary means of representing how changes in one variable propagate to downstream variables. Recent federated causal-learning methods recover global causal graphs from decentralized and heterogeneous data, with later studies considering scalability and nonidentical variable sets across clients \cite{yang2024fedcausal,guo2024fedcsl,wang2025federatedcausal}. Federated causal representations have also begun to support counterfactual reasoning in distributed industrial systems \cite{mohamed2026federatedcausal}. However, these methods mainly seek a single global graph or a shared latent causal representation and are not designed for fault diagnosis with overlapping operating mechanisms and incomplete local fault classes. SeaCausal-FL connects these directions by associating each uncertain operating mechanism with a local structural causal model, aligning corresponding mechanisms before aggregation, and conditioning class-specific parameter sharing on both mechanism activation and local fault evidence.

\section{The SeaCausal-FL Framework}

\subsection{System Model and Learning Objective}

As shown in Fig.~\ref{fig:framework}, SeaCausal-FL considers a federated maritime IoT system consisting of distributed maritime clients and a federated server. The main components are described as follows.

\begin{itemize}
\item \textbf{Maritime clients:} Let $\mathcal{K}=\{1,2,\ldots,K\}$ denote the set of participating clients. Each client stores a private collection of marine-engine sensor measurements and performs model training locally.

\item \textbf{Local sensor data:} The continuous sensor sequence at client $k$ is divided into fixed-length sliding windows. The local training dataset is denoted by
\begin{equation}
\mathcal{D}_{k}
=
\left\{
\left(
\mathbf{X}_{k,n},
\mathbf{M}_{k,n},
y_{k,n}
\right)
\right\}_{n=1}^{N_k},
\label{eq:local_dataset}
\end{equation}
where $N_k$ is the number of valid training windows at client $k$, $\mathbf{X}_{k,n}\in\mathbb{R}^{T\times D}$ is the $n$th sensor window with $T$ time steps and $D$ retained variables, and $\mathbf{M}_{k,n}\in\{0,1\}^{T\times D}$ is the corresponding missing-value mask, in which an entry of one indicates a missing observation. The label $y_{k,n}\in\mathcal{C}=\{0,\ldots,C-1\}$ identifies the fault class. A window is retained only when all of its observations belong to the same temporal partition and fault state, preventing temporal leakage and mixed-label supervision.

\item \textbf{Federated server:} The server initializes and broadcasts the global model, receives the aligned local parameters and compact evidence statistics, and performs evidence-aware model aggregation.
\end{itemize}

Let $P_k(\mathbf{X},\mathbf{M},Y)$ denote the local data distribution at client $k$. SeaCausal-FL does not impose a specific client-partition assumption. Under an IID partition,
$P_k(\mathbf{X},\mathbf{M},Y)=P_j(\mathbf{X},\mathbf{M},Y)$
for all client pairs. Under a heterogeneous partition, there exists at least one pair of clients satisfying
$P_k(\mathbf{X},\mathbf{M},Y)\neq P_j(\mathbf{X},\mathbf{M},Y)$
for $k\neq j$. Such heterogeneity is induced by operating-condition variation, label skew, or both. The same fault can produce different sensor responses under different loads, while some clients contain only a subset of the fault classes. In these cases, conventional sample-count-based aggregation does not distinguish how much evidence a client provides for a particular operating mechanism or fault class.

Let $\boldsymbol{\Theta}$ denote all trainable parameters of SeaCausal-FL, and let $\mathbf{p}_{k,n}(\boldsymbol{\Theta})\in[0,1]^C$ denote the predicted fault-probability vector for the $n$th window of client $k$. The local learning objective is defined as
\begin{equation}
\mathcal{F}_{k}(\boldsymbol{\Theta})
=
\frac{1}{N_k}
\sum_{n=1}^{N_k}
\ell_{\mathrm{cw}}
\left(
\mathbf{p}_{k,n}(\boldsymbol{\Theta}),
y_{k,n}
\right)
+
\mathcal{L}_{k}^{\mathrm{aux}}(\boldsymbol{\Theta}),
\label{eq:local_objective}
\end{equation}
where $\ell_{\mathrm{cw}}(\cdot)$ denotes the class-weighted cross-entropy loss. The auxiliary term $\mathcal{L}_{k}^{\mathrm{aux}}$ contains the objectives used for causal-branch supervision, fuzzy-mechanism regularization, structural-equation reconstruction, and causal-edge regularization. These objectives are introduced together with their corresponding modules in the following subsections.

Let $N=\sum_{k=1}^{K}N_k$ denote the total number of training windows across all clients. A reference sample-weighted federated objective is formulated as
\begin{equation}
\min_{\boldsymbol{\Theta}}
\mathcal{F}(\boldsymbol{\Theta})
=
\sum_{k=1}^{K}
\frac{N_k}{N}
\mathcal{F}_{k}(\boldsymbol{\Theta}).
\label{eq:global_objective}
\end{equation}
Equation~\eqref{eq:global_objective} describes the conventional client-level empirical weighting. SeaCausal-FL retains the coefficient $N_k/N$ only for shared non-class parameters rather than applying it uniformly to all parameter groups. Class-specific, mechanism-specific, and mechanism--class-specific parameters are aggregated according to local class counts, effective mechanism membership masses, and joint mechanism--class evidence, respectively.

\subsection{Overview of SeaCausal-FL}
\label{sec:overview}

Fig.~\ref{fig:framework} presents the overall architecture of SeaCausal-FL. At communication round $t$, the server broadcasts the current global model to a subset of maritime clients. Each selected client updates the model using its private sensor windows, aligns its locally learned mechanisms with the broadcast global reference, and uploads the aligned parameters together with compact evidence statistics for server aggregation. The framework contains three main components.

\textit{1) Client-Side Fuzzy Causal Diagnosis:} Each client first uses a shared temporal encoder to process the sensor window and its missing-value mask. The resulting representation is mapped to the base diagnostic logits as
\begin{equation}
\mathbf{z}^{\mathrm{base}}_{k,n}
=
g_{\boldsymbol{\theta}_{d}}
\left(
f_{\boldsymbol{\theta}_{e}}
\left(
\mathbf{X}_{k,n},
\mathbf{M}_{k,n}
\right)
\right),
\label{eq:base_logits}
\end{equation}
where $f_{\boldsymbol{\theta}_{e}}(\cdot)$ and $g_{\boldsymbol{\theta}_{d}}(\cdot)$ denote the shared temporal encoder and diagnostic head, respectively. This shared path learns fault characteristics that remain useful across different operating conditions.

To model operating-dependent physical responses, SeaCausal-FL extracts a physical state $\mathbf{s}_{k,n}$ from selected engine variables. Engine speed, water-brake load, and engine-room temperature form the operating context $\mathbf{c}_{k,n}$. An interval type-2 fuzzy layer assigns lower, midpoint, and upper weights, denoted by $\underline{\pi}_{k,n,r}$, $\pi_{k,n,r}$, and $\overline{\pi}_{k,n,r}$, respectively, to each latent operating mechanism $r\in\{1,\ldots,R\}$. The number of mechanisms is selected from the training operating contexts subject to a data-sufficiency constraint. The fuzzy centers, lower and upper scales, and mixture priors are initialized from the resulting data-driven prior and further optimized during local training. 

Each operating mechanism is associated with a structural causal model constructed within a physics-constrained candidate graph. Let $\mathbf{z}^{(r)}_{k,n}\in\mathbb{R}^{C}$ denote the mechanism-specific causal logits produced by mechanism $r$. The midpoint fuzzy weights combine these mechanism-specific outputs as
\begin{equation}
\mathbf{z}^{\mathrm{causal}}_{k,n}
=
\sum_{r=1}^{R}
\pi_{k,n,r}
\mathbf{z}^{(r)}_{k,n}.
\label{eq:causal_logits}
\end{equation}
The weighted causal output is introduced as a residual correction to the shared diagnostic logits:
\begin{equation}
\mathbf{z}_{k,n}
=
\mathbf{z}^{\mathrm{base}}_{k,n}
+
\lambda_{\mathrm{c}}
\mathbf{z}^{\mathrm{causal}}_{k,n},
\label{eq:fused_logits}
\end{equation}
where $\lambda_{\mathrm{c}}$ is a bounded learnable coefficient obtained from $\rho$ through a hyperbolic-tangent transformation. Since $\rho$ is initialized to zero, the shared diagnostic path determines the initial prediction, while the causal correction is introduced gradually during training.

\textit{2) Federated Mechanism Coordination:} Local mechanism indices do not necessarily share the same physical meaning across clients. Before upload, each selected client aligns its locally updated mechanisms with the broadcast global reference using operating-context centers and causal structures. If the preliminary assignment is nonidentity, label-independent intervention--response signatures are added to refine the matching. The aligned mechanism parameters and evidence statistics are then reordered before transmission.

The server applies evidence-aware aggregation to different parameter groups. Shared parameters use local sample counts, class-specific parameters use class counts, mechanism-specific fuzzy and causal parameters use effective membership masses, and mechanism--class-specific parameters use joint mechanism--class evidence. Clients with insufficient support therefore exert limited influence on the corresponding global parameters.

\textit{3) Interval Counterfactual Fault Reasoning:} The global model performs fault diagnosis using
$\mathbf{p}_{k,n}=\operatorname{softmax}(\mathbf{z}_{k,n})$
and supports interval counterfactual reasoning through abduction, action, and prediction. Given a factual physical state, SeaCausal-FL retains the mechanism-specific exogenous residuals, intervenes on an actionable variable, and propagates its effects through the downstream causal nodes. The factual temporal logits and mechanism assignment remain fixed. The midpoint mechanism weights produce the nominal counterfactual risk, while the feasible IT2 weight set defined by the lower and upper bounds determines its interval.

The complete training procedure is summarized in Algorithm~\ref{alg:seacausal_training}. In each communication round, selected clients jointly train the shared diagnostic and fuzzy causal components, collect mechanism-related evidence, and align their local mechanisms before upload. The server then aggregates different parameter groups using the corresponding sample, class, mechanism, and mechanism--class evidence.

\begin{algorithm}[!t]
\caption{Training Procedure of SeaCausal-FL}
\label{alg:seacausal_training}
\footnotesize
\begin{algorithmic}[1]

\Require Local datasets $\{\mathcal{D}_{k}\}_{k=1}^{K}$; data-driven prior $\mathcal{P}$; intervention grid $\mathcal{I}$; communication rounds $T_{\mathrm{FL}}$; local epochs $E$; participation ratio $q$; class weights $\boldsymbol{\omega}$; patience $H$.
\Ensure Best global model $\boldsymbol{\Theta}^{\star}$.

\State Initialize $\boldsymbol{\Theta}^{0}$ from $\mathcal{P}$; set $s^{\star}\gets-\infty$ and $p\gets0$.

\For{$t=0,1,\ldots,T_{\mathrm{FL}}-1$}
    \State Select $\mathcal{S}_{t}\subseteq\mathcal{K}$ according to $q$ and broadcast $\boldsymbol{\Theta}^{t}$.

    \Statex \textit{Client-side learning and mechanism alignment}
    \ForAll{$k\in\mathcal{S}_{t}$ \textbf{in parallel}}
        \State Set $\boldsymbol{\Theta}_{k}\gets\boldsymbol{\Theta}^{t}$ and initialize $m_{k,r}^{t}$, $m_{k,r,c}^{t}$, and $n_{k,c}^{t}$ to zero.

        \For{$e=1,2,\ldots,E$}
            \ForAll{minibatches $\mathcal{B}\subset\mathcal{D}_{k}$}
                \State Compute the shared, fuzzy, causal, and fused outputs.
                \State Compute $\mathcal{L}_{k}$ using \eqref{eq:complete_client_loss} and update $\boldsymbol{\Theta}_{k}$ using AdamW with gradient clipping.
                \State Accumulate $m_{k,r}^{t}$, $m_{k,r,c}^{t}$, and $n_{k,c}^{t}$ according to \eqref{eq:local_evidence_statistics}.
            \EndFor
        \EndFor

        \State Set $\boldsymbol{\Theta}_{k}^{t+1}\gets\boldsymbol{\Theta}_{k}$ and collect
        $\mathcal{E}_{k}^{t}\gets\{N_k,m_{k,r}^{t},m_{k,r,c}^{t},n_{k,c}^{t}\}$.
        \State Construct $\mathbf{C}_{k}^{(0)}$ using \eqref{eq:prealignment_cost} and obtain $\sigma_{k}^{(0)}$ by Hungarian matching.

        \If{$\sigma_{k}^{(0)}$ is not the identity assignment}
            \State Compute local and reference intervention--response signatures on the same private calibration minibatches.
            \State Construct $\mathbf{C}_{k}$ using \eqref{eq:complete_alignment_cost} and obtain $\sigma_{k}^{t}$ by Hungarian matching.
        \Else
            \State Set $\sigma_{k}^{t}\gets\sigma_{k}^{(0)}$.
        \EndIf

        \State Jointly reorder $\boldsymbol{\Theta}_{k}^{t+1}$ and $\mathcal{E}_{k}^{t}$ according to $\sigma_{k}^{t}$.
        \State Upload the aligned local parameters and evidence statistics.
    \EndFor

    \Statex \textit{Evidence-aware server aggregation}
    \State Aggregate the four parameter groups using $N_k$, $n_{k,c}^{t}$, $m_{k,r}^{t}$, and $m_{k,r,c}^{t}$ according to \eqref{eq:evidence_aggregation_operator}.
    \State Retain the previous global parameter block when its total evidence is zero.
    \State Obtain $\boldsymbol{\Theta}^{t+1}$ and its validation F1-score $s^{t+1}$.

    \If{$s^{t+1}>s^{\star}$}
        \State $\boldsymbol{\Theta}^{\star}\gets\boldsymbol{\Theta}^{t+1}$, $s^{\star}\gets s^{t+1}$, and $p\gets0$.
    \Else
        \State $p\gets p+1$.
    \EndIf

    \If{$p\geq H$}
        \State \textbf{break}
    \EndIf
\EndFor

\State \Return $\boldsymbol{\Theta}^{\star}$.

\end{algorithmic}
\end{algorithm}

\subsection{Client-Side Fuzzy Causal Diagnosis}
\label{sec:client_diagnosis}

At each selected client, SeaCausal-FL jointly learns a shared temporal diagnostic path and an operating-dependent fuzzy causal path. The shared path captures fault characteristics that recur across clients, while the causal path models physical relations that vary with operating conditions. Instead of training an independent classifier for every mechanism, SeaCausal-FL retains a shared diagnostic backbone and introduces lightweight mechanism-conditioned causal corrections.

\subsubsection{Shared Temporal Diagnostic Path}
\label{sec:shared_temporal_path}

\begin{figure}[!t]
\centering
\includegraphics[width=0.75\columnwidth]{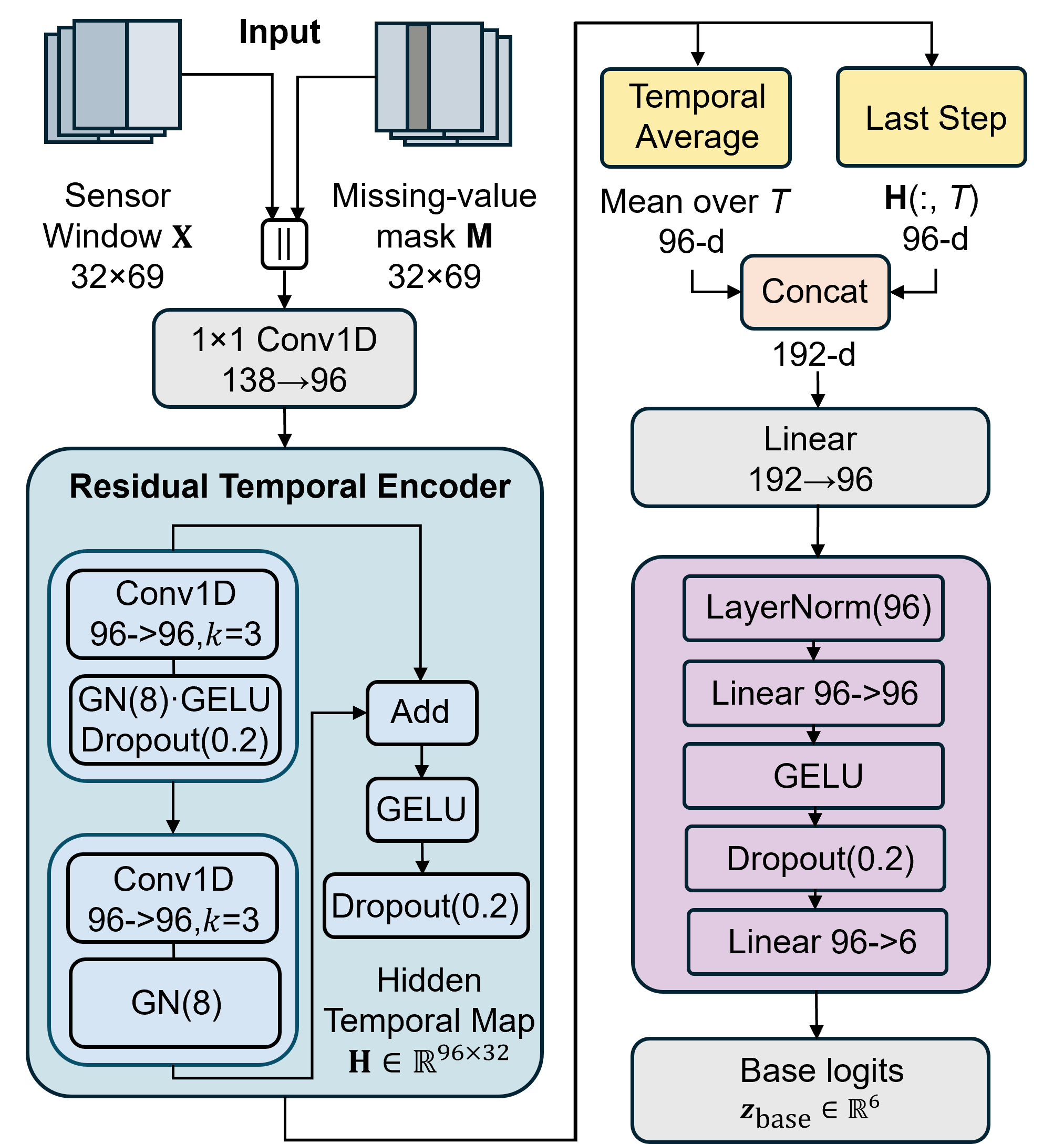}
\caption{Detailed architecture of the shared temporal diagnostic path.}
\label{fig:diagnostic_structure}
\end{figure}

The shared path provides a stable diagnostic basis before the operating-dependent causal correction becomes active. Since missing observations may affect the reliability of temporal features, the sensor window and its missing-value mask are concatenated along the feature dimension. The resulting tensor is processed by a two-layer one-dimensional residual convolutional encoder. Each convolutional layer is followed by group normalization, while GELU activation and dropout are applied within the residual path. Unlike batch normalization, group normalization does not rely on client-specific batch statistics and is therefore suitable for heterogeneous federated clients. The detailed architecture of this shared temporal diagnostic path is illustrated in Fig.~\ref{fig:diagnostic_structure}.

Let $\mathbf{H}_{k,n}\in\mathbb{R}^{d_h\times T}$ denote the hidden temporal feature map produced from $(\mathbf{X}_{k,n},\mathbf{M}_{k,n})$, where $d_h$ is the hidden-channel dimension. SeaCausal-FL combines the average temporal response and the last-step feature as
\begin{equation}
\mathbf{h}_{k,n}
=
\mathbf{W}_{p}
\left[
\operatorname{Avg}_{\tau}
\left(
\mathbf{H}_{k,n}
\right)
\mathbin{\|}
\mathbf{H}_{k,n}(:,T)
\right]
+
\mathbf{b}_{p},
\label{eq:shared_temporal_representation}
\end{equation}
where $\mathbin{\|}$ denotes vector concatenation. The average summary describes the overall behavior of the sensor window, while the last-step feature retains its most recent temporal response. The representation $\mathbf{h}_{k,n}$ is then mapped to the base diagnostic logits through the diagnostic head defined in \eqref{eq:base_logits}. This shared path uses all retained sensor variables and provides the primary fault-classification signal during local training.

\subsubsection{Data-Driven Interval Type-2 Fuzzy Mechanism Modeling}
\label{sec:data_driven_it2}

Fixed low-, medium-, and high-load thresholds are inappropriate when maritime operating conditions vary continuously and overlap. SeaCausal-FL therefore derives latent operating mechanisms from training data and represents uncertainty in their assignments using interval type-2 fuzzy memberships.

For each physical variable, the most recent valid value in the current sensor window is retained to construct the physical state $\mathbf{s}_{k,n}$. If a physical variable has no valid observation in the window, its value is set to zero and its own structural reconstruction term is masked out. Engine speed, water-brake load, and engine-room temperature are selected as exogenous operating variables, giving
\begin{equation}
\mathbf{c}_{k,n}
=
\mathbf{P}_{c}\mathbf{s}_{k,n},
\label{eq:fuzzy_operating_context}
\end{equation}
where $\mathbf{P}_{c}$ is a fixed operating-context selection matrix. The membership function depends only on these exogenous operating variables; fault labels and downstream physical responses are not included in $\mathbf{c}_{k,n}$.

To estimate the data-driven prior, a diagonal-covariance Gaussian mixture model (GMM) is fitted to the operating contexts extracted from normal training windows when the number of normal windows satisfies the minimum prior-estimation requirement $N_{\mathrm{prior}}^{\min}$; otherwise, all training contexts are used. Candidate mechanism numbers are considered within $\mathcal{R}=\{1,\ldots,R_{\max}\}$.

For an $R$-component GMM, let $\gamma_{n,r}^{(R)}$ denote the posterior responsibility of mechanism $r$ for training context $n$. Its soft effective sample count is defined as
\begin{equation}
N_{r}^{\mathrm{soft}}(R)
=
\sum_{n}
\gamma_{n,r}^{(R)}.
\label{eq:soft_mechanism_count}
\end{equation}
The corresponding hard-assignment count is
\begin{equation}
N_{r}^{\mathrm{hard}}(R)
=
\sum_{n}
\mathbb{I}
\left\{
r=
\arg\max_{j}
\gamma_{n,j}^{(R)}
\right\}.
\label{eq:hard_mechanism_count}
\end{equation}

A candidate $R$ is considered eligible only when every mechanism has sufficient support. Specifically,
\begin{equation}
\min_{r}
N_{r}^{\mathrm{soft}}(R)
\geq
N_{\mathrm{soft}}^{\min},
\label{eq:soft_sufficiency}
\end{equation}
and
\begin{equation}
\min_{r}
N_{r}^{\mathrm{hard}}(R)
\geq
N_{\mathrm{hard}}^{\min}.
\label{eq:hard_sufficiency}
\end{equation}
The candidates satisfying both conditions form the eligible set $\mathcal{R}_{\mathrm{elig}}$.

Among these candidates, the mechanism number is selected by
\begin{equation}
R^{\star}
=
\arg\min_{R\in\mathcal{R}_{\mathrm{elig}}}
\left(
\operatorname{BIC}_{R}
+
2\mathcal{H}_{R}
\right),
\label{eq:mechanism_number_selection}
\end{equation}
where the posterior assignment entropy is
\begin{equation}
\mathcal{H}_{R}
=
-
\sum_{n}
\sum_{r=1}^{R}
\gamma_{n,r}^{(R)}
\log
\gamma_{n,r}^{(R)}.
\label{eq:mechanism_assignment_entropy}
\end{equation}
The BIC term favors a parsimonious fit to the operating-context distribution, while the entropy term penalizes ambiguous mechanism assignments. The number of fuzzy causal mechanisms used in subsequent training is set to $R=R^{\star}$.

The resulting GMM centers, scales, and mixture probabilities initialize the trainable fuzzy parameters. The lower and upper scales account for finite-sample uncertainty in both scale and center estimation rather than using a manually chosen footprint width. Let $\boldsymbol{\xi}_{r}$ denote the center of mechanism $r$. For $a\in\{\mathrm{L},\mathrm{U}\}$, the membership candidate obtained from scale $\boldsymbol{\sigma}_{r}^{a}$ is
\begin{equation}
\widetilde{\mu}_{k,n,r}^{a}
=
\exp
\left[
-\frac{1}{2}
\left\|
\frac{
\mathbf{c}_{k,n}-\boldsymbol{\xi}_{r}
}{
\boldsymbol{\sigma}_{r}^{a}
}
\right\|_{2}^{2}
\right].
\label{eq:interval_membership}
\end{equation}
The lower and upper memberships are defined as
$\underline{\mu}_{k,n,r}
=\min\{\widetilde{\mu}_{k,n,r}^{\mathrm{L}},
\widetilde{\mu}_{k,n,r}^{\mathrm{U}}\}$
and
$\overline{\mu}_{k,n,r}
=\max\{\widetilde{\mu}_{k,n,r}^{\mathrm{L}},
\widetilde{\mu}_{k,n,r}^{\mathrm{U}}\}$,
respectively.

Let $\alpha_r$ denote the trainable mixture probability initialized from the GMM, with $\alpha_r\geq0$ and $\sum_{r=1}^{R}\alpha_r=1$. The midpoint mechanism weight is
\begin{equation}
\pi_{k,n,r}
=
\frac{
\alpha_r
\left(
\underline{\mu}_{k,n,r}
+
\overline{\mu}_{k,n,r}
\right)
}{
\displaystyle
\sum_{j=1}^{R}
\alpha_j
\left(
\underline{\mu}_{k,n,j}
+
\overline{\mu}_{k,n,j}
\right)
+
\varepsilon
}.
\label{eq:midpoint_mechanism_weight}
\end{equation}
The interval memberships define a normalized uncertainty half-width around the midpoint mechanism weight. For mechanism $r$, it is computed as
\begin{equation}
\delta_{k,n,r}
=
\frac{
\frac{1}{2}\alpha_r
\left(
\overline{\mu}_{k,n,r}
-
\underline{\mu}_{k,n,r}
\right)
}{
\displaystyle
\frac{1}{2}
\sum_{j=1}^{R}
\alpha_j
\left(
\underline{\mu}_{k,n,j}
+
\overline{\mu}_{k,n,j}
\right)
+
\varepsilon
}.
\label{eq:mechanism_weight_halfwidth}
\end{equation}
Collecting these values gives
$\boldsymbol{\delta}_{k,n}
=
[\delta_{k,n,1},\ldots,\delta_{k,n,R}]^{\top}$.
The provisional lower and upper bounds are then constructed separately as
\begin{equation}
\mathbf{l}_{k,n}
=
\left[
\boldsymbol{\pi}_{k,n}
-
\boldsymbol{\delta}_{k,n}
\right]_{[0,1]},
\label{eq:provisional_lower_bound}
\end{equation}
and
\begin{equation}
\mathbf{u}_{k,n}
=
\left[
\boldsymbol{\pi}_{k,n}
+
\boldsymbol{\delta}_{k,n}
\right]_{[0,1]},
\label{eq:provisional_upper_bound}
\end{equation}
where $[\cdot]_{[0,1]}$ denotes element-wise clipping. If
$\mathbf{1}^{\top}\mathbf{l}_{k,n}>1$, the lower bound is normalized
by its sum. If $\mathbf{1}^{\top}\mathbf{u}_{k,n}<1$, the missing
probability mass is distributed over the remaining upper-bound
capacity. The bounds are finally adjusted to enclose the midpoint
weight, yielding
\begin{equation}
\underline{\boldsymbol{\pi}}_{k,n}
\preceq
\boldsymbol{\pi}_{k,n}
\preceq
\overline{\boldsymbol{\pi}}_{k,n},
\label{eq:fuzzy_interval_enclosure}
\end{equation}
with
\begin{equation}
\mathbf{1}^{\top}
\underline{\boldsymbol{\pi}}_{k,n}
\leq
1
\leq
\mathbf{1}^{\top}
\overline{\boldsymbol{\pi}}_{k,n}.
\label{eq:fuzzy_interval_feasibility}
\end{equation}
Thus, the interval contains at least one valid normalized
mechanism-weight vector.

During local training, three auxiliary terms constrain the fuzzy
mechanism model. Let $\boldsymbol{\xi}_{r}^{0}$,
$\boldsymbol{\sigma}_{r}^{L,0}$,
$\boldsymbol{\sigma}_{r}^{U,0}$, and $\alpha_{r}^{0}$ denote the
GMM-derived prior parameters. The prior uncertainty scale is defined
as
$\boldsymbol{\tau}_{r}
=
(\boldsymbol{\sigma}_{r}^{U,0}
-
\boldsymbol{\sigma}_{r}^{L,0})/2$.
The context negative log-likelihood is

\begin{equation}
\resizebox{0.94\linewidth}{!}{$
\displaystyle
\mathcal{L}_{k}^{\mathrm{context}}
=
-\frac{1}{|\mathcal{B}|}
\sum_{n\in\mathcal{B}}
\log
\sum_{r=1}^{R}
\frac{\alpha_r}
{\prod_d \sigma_{r,d}^{L}}
\exp
\left[
-\frac{1}{2}
\left\|
\frac{
\mathbf{c}_{k,n}
-
\boldsymbol{\xi}_r
}{
\boldsymbol{\sigma}_{r}^{L}
}
\right\|_{2}^{2}
\right]
$}
\label{eq:context_nll}
\end{equation}

The normalized prior-drift loss is

\begin{equation}
\resizebox{0.94\linewidth}{!}{$
\displaystyle
\mathcal{L}_{k}^{\mathrm{drift}}
=
\operatorname{Mean}
\left[
\left(
\frac{
\boldsymbol{\xi}
-
\boldsymbol{\xi}^{0}
}{
\boldsymbol{\tau}
}
\right)^{2}
+
\left(
\frac{
\boldsymbol{\sigma}^{L}
-
\boldsymbol{\sigma}^{L,0}
}{
\boldsymbol{\tau}
}
\right)^{2}
+
\left(
\frac{
\boldsymbol{\sigma}^{U}
-
\boldsymbol{\sigma}^{U,0}
}{
\boldsymbol{\tau}
}
\right)^{2}
\right]
$}
\label{eq:fuzzy_prior_drift}
\end{equation}

Deviations of the trainable mixture probabilities from the GMM
prior are controlled by
\begin{equation}
\mathcal{L}_{k}^{\mathrm{mix}}
=
\sum_{r=1}^{R}
\alpha_r
\log
\frac{\alpha_r}{\alpha_r^{0}}.
\label{eq:mixture_prior_kl}
\end{equation}
These terms permit local adaptation while constraining the fuzzy
mechanisms to remain close to their data-derived operating meanings.

\subsubsection{Physics-Constrained Causal Mechanism Bank}
\label{sec:causal_mechanism_bank}

Learning an unrestricted causal graph from a small and non-IID local dataset may produce cyclic or physically implausible relations. SeaCausal-FL therefore defines an auditable candidate graph according to the ordered engine processes of control, air intake, cooling, combustion, exhaust, and shaft power. The graph specifies admissible causal directions and expected coefficient signs, while edge probabilities and coefficient magnitudes remain trainable.

Let $\mathcal{E}_{\mathrm{phy}}$ denote the physics-constrained candidate edge set, and let $\operatorname{Pa}(j)$ denote the candidate parents of physical node $j$. For mechanism $r$, the edge probability associated with $(i,j)\in\mathcal{E}_{\mathrm{phy}}$ is
$g_{ij}^{(r)}=\operatorname{sigmoid}(\eta_{ij}^{(r)})$.
The corresponding sign-constrained linear and saturating coefficients are denoted by $a_{ij}^{(r)}$ and $c_{ij}^{(r)}$, respectively. The structural prediction of node $j$ is
\begin{equation}
\widehat{s}_{k,n,j}^{(r)}
=
b_{j}^{(r)}
+
\sum_{i\in\operatorname{Pa}(j)}
g_{ij}^{(r)}
\left[
a_{ij}^{(r)}s_{k,n,i}
+
c_{ij}^{(r)}
\tanh
\left(
s_{k,n,i}
\right)
\right].
\label{eq:mechanism_structural_equation}
\end{equation}
The linear term represents the local response around an operating point, while the saturating term describes bounded nonlinear effects without introducing an unrestricted neural structural equation. The tiered candidate graph ensures that all admissible edges follow the predefined physical ordering and therefore remain acyclic.

Let $o_{k,n,j}\in\{0,1\}$ indicate whether node $j$ is observed in the current window, and let $\chi_j\in\{0,1\}$ indicate whether node $j$ has at least one admissible parent and is therefore reconstructable. The effective structural weight is
$\omega_{k,n,j,r}=\pi_{k,n,r}o_{k,n,j}\chi_j$.
For a local minibatch $\mathcal{B}$, the mechanism-weighted structural loss is
\begin{equation}
\mathcal{L}_{k}^{\mathrm{sem}}
=
\frac{
\displaystyle
\sum_{n\in\mathcal{B}}
\sum_{r=1}^{R}
\sum_{j=1}^{J}
\omega_{k,n,j,r}
\left(
\widehat{s}_{k,n,j}^{(r)}
-
s_{k,n,j}
\right)^{2}
}{
\displaystyle
\sum_{n\in\mathcal{B}}
\sum_{r=1}^{R}
\sum_{j=1}^{J}
\omega_{k,n,j,r}
+
\varepsilon
},
\label{eq:weighted_sem_loss}
\end{equation}
where $J$ is the number of retained physical nodes. Thus, a mechanism is trained primarily by samples with high membership support, while unobserved target nodes and nonreconstructable nodes do not contribute to its structural loss.

Before federated training, the edge probabilities and structural coefficients are initialized using sign-constrained robust regressions on the training data. The ridge regularization strength is selected by generalized cross-validation, while comparisons between reduced and complete structural equations provide BIC-based edge evidence. Let $g_{ij}^{0,(r)}$ denote the resulting edge-probability prior. Deviations from this prior are controlled by
{\small
\begin{equation}
\mathcal{L}_{k}^{\mathrm{edge}}
=
\frac{1}{R|\mathcal{E}_{\mathrm{phy}}|}
\sum_{r=1}^{R}
\sum_{(i,j)\in\mathcal{E}_{\mathrm{phy}}}
\operatorname{KL}
\left[
\operatorname{Bern}
\left(
g_{ij}^{(r)}
\right)
\middle\|
\operatorname{Bern}
\left(
g_{ij}^{0,(r)}
\right)
\right].
\label{eq:edge_prior_kl}
\end{equation}
}
This term discourages unsupported changes to the data-driven edge prior while retaining the ability to adapt edge probabilities when sufficient local evidence is available.

\subsubsection{Causal Residual Fault Prediction}
\label{sec:causal_residual_prediction}

The physical state is interpretable but contains less diagnostic information than the complete temporal sensor window. SeaCausal-FL therefore does not replace the shared temporal classifier with the causal branch. Instead, all mechanisms share a nonlinear physical-state outcome network, while each mechanism learns only a zero-initialized linear residual.

Let $\mathbf{q}(\mathbf{s}_{k,n})\in\mathbb{R}^{C}$ denote the logits produced by the shared causal outcome network. The causal logits associated with mechanism $r$ are
\begin{equation}
\mathbf{z}_{k,n}^{(r)}
=
\mathbf{q}
\left(
\mathbf{s}_{k,n}
\right)
+
\mathbf{W}_{r}^{\mathrm{res}}
\mathbf{s}_{k,n}
+
\mathbf{b}_{r}^{\mathrm{res}}.
\label{eq:mechanism_specific_causal_logits}
\end{equation}
The residual parameters $\mathbf{W}_{r}^{\mathrm{res}}$ and $\mathbf{b}_{r}^{\mathrm{res}}$ are initialized to zero. Mechanism specialization is therefore introduced gradually during local optimization rather than imposed at initialization. The mechanism-specific outputs are combined using \eqref{eq:causal_logits} and added to the shared diagnostic logits through the zero-initialized bounded scale in \eqref{eq:fused_logits}.

Since the causal contribution to the fused logits is initially zero, the causal branch receives an independent auxiliary classification signal:
\begin{equation}
\mathcal{L}_{k}^{\mathrm{causal}}
=
\frac{1}{|\mathcal{B}|}
\sum_{n\in\mathcal{B}}
\ell_{\mathrm{cw,ls}}
\left(
\mathbf{z}_{k,n}^{\mathrm{causal}},
y_{k,n}
\right),
\label{eq:causal_auxiliary_loss}
\end{equation}
where $\ell_{\mathrm{cw,ls}}(\cdot)$ denotes the class-weighted cross-entropy computed from logits with the same label-smoothing setting as the primary classification loss.

Fault classification remains the primary objective under class imbalance. Let
$\mathcal{A}=\{\mathrm{causal},\mathrm{sem},\mathrm{edge},\mathrm{context},\mathrm{drift},\mathrm{mix}\}$ denote the active auxiliary-loss set, corresponding to causal classification, structural reconstruction, edge-prior regularization, context negative log-likelihood, fuzzy-prior drift, and mixture-prior KL divergence, respectively. Their relative contributions are automatically learned through log-variance parameters:
\begin{equation}
\mathcal{L}_{k}
=
\mathcal{L}_{k}^{\mathrm{cls}}
+
\frac{1}{|\mathcal{A}|}
\sum_{m\in\mathcal{A}}
\left[
\frac{1}{2}
\exp(-v_m)
\mathcal{L}_{k,m}
+
\frac{1}{2}v_m
\right],
\label{eq:complete_client_loss}
\end{equation}
where $v_m$ is a trainable log-variance parameter constrained to a finite numerical range during optimization. The primary loss $\mathcal{L}_{k}^{\mathrm{cls}}$ is computed from the fused logits $\mathbf{z}_{k,n}$ and remains outside the automatic loss balancer. Its coefficient is therefore fixed at one and cannot be reduced by the uncertainty-based balancing mechanism. This design avoids manually assigning a separate coefficient to every fuzzy and causal auxiliary objective.

\subsection{Federated Mechanism Coordination}
\label{sec:federated_coordination}

After local training, mechanisms with the same index may no longer represent the same operating condition at different clients. Directly averaging such parameters can mix fuzzy prototypes and causal structures with different physical meanings. Before uploading its locally updated model, each selected client therefore aligns its mechanisms with the broadcast global reference. The aligned parameters and mechanism-related evidence statistics are subsequently used for server aggregation. The coordination procedure consists of context--structure prealignment, conditional intervention--response refinement, and parameter-group-specific aggregation.

\subsubsection{Context--Structure Prealignment}
\label{sec:context_structure_alignment}

The local model is initialized from the current global model at the beginning of each communication round. Most mechanisms therefore retain their original ordering after local optimization. Computing intervention responses for every client in every round would therefore introduce unnecessary computational overhead. SeaCausal-FL first performs a low-cost prealignment using the operating-context centers and causal structures already contained in the local and global models.

Let $\boldsymbol{\xi}_{k,r}^{t+1}$ denote the center of local mechanism $r$ after client $k$ completes its local update in round $t$, and let $\boldsymbol{\xi}_{s}^{t}$ denote the center of global mechanism $s$. The context-center distance is
\begin{equation}
D_{k,r,s}^{\mathrm{ctx}}
=
\left\|
\boldsymbol{\xi}_{k,r}^{t+1}
-
\boldsymbol{\xi}_{s}^{t}
\right\|_{2}.
\label{eq:context_alignment_cost}
\end{equation}

Let $\mathbf{B}_{k,r}^{t+1}\in\mathbb{R}^{J\times J}$ and $\mathbf{B}_{s}^{t}\in\mathbb{R}^{J\times J}$ denote the effective causal coefficient matrices of the local and global mechanisms, respectively. For an admissible edge $(i,j)\in\mathcal{E}_{\mathrm{phy}}$, the corresponding entry is defined as $B_{ij}^{(r)}=g_{ij}^{(r)} \left(a_{ij}^{(r)}+c_{ij}^{(r)}\right)$ and is set to zero otherwise. This matrix provides a compact summary of the edge probability and the linear and saturating structural coefficients. The structure distance is defined as
\begin{equation}
D_{k,r,s}^{\mathrm{str}}
=
\frac{1}{J}
\left\|
\mathbf{B}_{k,r}^{t+1}
-
\mathbf{B}_{s}^{t}
\right\|_{\mathrm{F}},
\label{eq:structure_alignment_cost}
\end{equation}
where $J$ is the number of physical variables and $\|\cdot\|_{\mathrm{F}}$ denotes the Frobenius norm.

Since the two distance matrices may have different numerical scales, SeaCausal-FL applies positive-median scaling:
\begin{equation}
\operatorname{RScale}(\mathbf{D})
=
\frac{\mathbf{D}}
{
\operatorname{median}
\left\{
D_{r,s}:D_{r,s}>0
\right\}
+
\varepsilon
}.
\label{eq:robust_cost_scaling}
\end{equation}
When no positive entry exists, the scale is set to one. The low-cost prealignment matrix is then
\begin{equation}
\mathbf{C}_{k}^{(0)}
=
\frac{1}{2}
\left[
\operatorname{RScale}
\left(
\mathbf{D}_{k}^{\mathrm{ctx}}
\right)
+
\operatorname{RScale}
\left(
\mathbf{D}_{k}^{\mathrm{str}}
\right)
\right].
\label{eq:prealignment_cost}
\end{equation}
This operation removes the scale difference between the context and structure distances, while their contributions remain equally weighted.

Let $\mathfrak{S}_{R}$ denote the set of permutations of $R$ mechanisms. The preliminary assignment is obtained from
\begin{equation}
\sigma_{k}^{(0)}
=
\arg\min_{\sigma\in\mathfrak{S}_{R}}
\sum_{s=1}^{R}
C_{k,\sigma(s),s}^{(0)},
\label{eq:prealignment_assignment}
\end{equation}
where $\sigma_{k}^{(0)}(s)$ is the local mechanism assigned to global mechanism $s$. The assignment is solved using the Hungarian algorithm. If $\sigma_{k}^{(0)}$ is the identity permutation, the local ordering is retained and the intervention-based refinement is skipped.

\subsubsection{Intervention--Response Signature Matching}
\label{sec:response_signature_alignment}

Similar fuzzy centers and causal coefficient matrices do not necessarily imply equivalent physical behavior. Small differences distributed across several causal paths can produce different downstream responses after intervention. SeaCausal-FL therefore evaluates intervention--response signatures only when the context--structure prealignment indicates a possible permutation.

The intervention grid $\mathcal{I}$ contains feasible values of actionable physical variables. These values are obtained from client-level empirical quantiles of the training windows and consolidated across clients without transmitting raw sensor observations. For each intervention $(a,v)\in\mathcal{I}$, physical variable $a$ is assigned value $v$. The factual mechanism-specific exogenous residuals are retained, and the intervention effects are propagated through the downstream physical nodes.

Let $\{\mathcal{B}_{k,b}^{\mathrm{cal}}\}_{b=1}^{B_k}$ denote the private calibration minibatches used for mechanism matching, where $B_k$ is bounded by the configured signature-batch budget. For $u\in\{\mathrm{loc},\mathrm{ref}\}$, representing the locally updated model and the broadcast global reference model, respectively, define the robust response of mechanism $r$ under intervention $(a,v)$ as
\begin{equation}
\boldsymbol{\delta}_{k,r}^{u}(a,v)
=
\frac{1}{B_k}
\sum_{b=1}^{B_k}
\operatorname{median}_{n\in\mathcal{B}_{k,b}^{\mathrm{cal}}}
\left[
\mathbf{s}_{k,n}^{\mathrm{cf},u,r}(a\leftarrow v)
-
\mathbf{s}_{k,n}
\right].
\label{eq:robust_intervention_response}
\end{equation}
The intervention--response signature is then obtained by concatenating the responses over all interventions:
\begin{equation}
\boldsymbol{\psi}_{k,r}^{u}
=
\operatorname{vec}_{(a,v)\in\mathcal{I}}
\left[
\boldsymbol{\delta}_{k,r}^{u}(a,v)
\right].
\label{eq:intervention_response_signature}
\end{equation}

Thus, each signature concatenates the minibatch-averaged median physical-state changes produced by all interventions. The locally updated model and the global reference are evaluated on the same private calibration minibatches, ensuring that their response signatures are comparable. The signature does not use fault labels and is therefore valid even when a client contains only a subset of the fault classes.

The intervention--response distance between local mechanism $r$ and global mechanism $s$ is
\begin{equation}
D_{k,r,s}^{\mathrm{resp}}
=
\left\|
\boldsymbol{\psi}_{k,r}^{\mathrm{loc}}
-
\boldsymbol{\psi}_{k,s}^{\mathrm{ref}}
\right\|_{2}.
\label{eq:response_alignment_cost}
\end{equation}
For the complete SeaCausal-FL model, the final alignment cost is
\begin{equation}
\mathbf{C}_{k}
=
\frac{1}{3}
\left[
\operatorname{RScale}
\left(
\mathbf{D}_{k}^{\mathrm{ctx}}
\right)
+
\operatorname{RScale}
\left(
\mathbf{D}_{k}^{\mathrm{str}}
\right)
+
\operatorname{RScale}
\left(
\mathbf{D}_{k}^{\mathrm{resp}}
\right)
\right].
\label{eq:complete_alignment_cost}
\end{equation}
The final assignment is obtained as
\begin{equation}
\sigma_{k}^{t}
=
\arg\min_{\sigma\in\mathfrak{S}_{R}}
\sum_{s=1}^{R}
C_{k,\sigma(s),s}.
\label{eq:complete_alignment_assignment}
\end{equation}

The fuzzy parameters, mechanism-indexed fuzzy priors, causal parameters, mechanism-specific output parameters, and mechanism-related evidence statistics are reordered according to $\sigma_{k}^{t}$. Shared parameters and class-count statistics are not permuted. Only the aligned model parameters and compact evidence summaries are used for aggregation; raw calibration windows and sample-level fault labels remain at the client.

\subsubsection{Server-side Evidence-Aware Aggregation}
\label{sec:evidence_aware_aggregation}

Conventional sample-count averaging assumes that every local sample provides equal evidence for every model parameter. This assumption is unsuitable for SeaCausal-FL because a client may strongly activate only a subset of operating mechanisms and may contain no samples from some fault classes. The server therefore selects the aggregation weight according to the parameter group being updated.

During local training, the midpoint fuzzy weights and fault labels already available in each forward pass are used to construct three complementary evidence statistics without an additional traversal of the local dataset. Let $\mathcal{T}_{k}^{t}$ denote the multiset of local sample occurrences processed by client $k$ during communication round $t$, including their repeated occurrences across local epochs. For mechanism $r$ and fault class $c$, the mechanism, mechanism--class, and class evidence are defined as
\begin{equation}
\begin{aligned}
m_{k,r}^{t}
&=
\sum_{n\in\mathcal{T}_{k}^{t}}
\pi_{k,n,r},
\\
m_{k,r,c}^{t}
&=
\sum_{n\in\mathcal{T}_{k}^{t}}
\pi_{k,n,r}
\mathbb{I}
\left\{
y_{k,n}=c
\right\},
\\
n_{k,c}^{t}
&=
\sum_{n\in\mathcal{T}_{k}^{t}}
\mathbb{I}
\left\{
y_{k,n}=c
\right\}.
\end{aligned}
\label{eq:local_evidence_statistics}
\end{equation}
Here, $m_{k,r}^{t}$ measures the effective support for mechanism $r$, $m_{k,r,c}^{t}$ measures the class-$c$ evidence observed under that mechanism, and $n_{k,c}^{t}$ records the overall evidence for class $c$. The mechanism-indexed statistics $m_{k,r}^{t}$ and $m_{k,r,c}^{t}$ are reordered together with the corresponding mechanism parameters after alignment, whereas $n_{k,c}^{t}$ is unaffected by the permutation.

Let $\widetilde{\boldsymbol{\vartheta}}_{k}^{t+1}$ denote an aligned local parameter block and let $w_k$ denote its corresponding evidence weight. The server aggregation operator is
\begin{equation}
\begin{split}
&\operatorname{Agg}
\left(
\left\{
\widetilde{\boldsymbol{\vartheta}}_{k}^{t+1},
w_k
\right\}_{k\in\mathcal{S}_{t}};
\boldsymbol{\vartheta}^{t}
\right)
\\[-1mm]
&\quad =
\begin{cases}
\displaystyle
\sum_{k\in\mathcal{S}_{t}}
\frac{w_k}
{\sum_{j\in\mathcal{S}_{t}} w_j}
\widetilde{\boldsymbol{\vartheta}}_{k}^{t+1},
&
\displaystyle
\sum_{j\in\mathcal{S}_{t}} w_j > 0,
\\[3mm]
\boldsymbol{\vartheta}^{t},
&
\displaystyle
\sum_{j\in\mathcal{S}_{t}} w_j = 0.
\end{cases}
\end{split}
\label{eq:evidence_aggregation_operator}
\end{equation}
Thus, when no participating client provides evidence for a parameter block, the corresponding parameter from the previous global model is retained.

Shared non-class parameters, including the temporal encoder, shared hidden layers, and other common parameters, use $w_k=N_k$. The class-$c$ rows of the diagnostic output layer and the shared causal output classifier use $w_k=n_{k,c}^{t}$. The trainable parameters associated with fuzzy mechanism $r$ and its structural causal model use $w_k=m_{k,r}^{t}$. Finally, the class-$c$ output parameters specific to mechanism $r$ use $w_k=m_{k,r,c}^{t}$.

The resulting global model exploits the evidence available from all participating clients without allowing a client with weak mechanism activation or a missing fault class to influence an unsupported parameter group. This coordination preserves the correspondence and evidence support of the fuzzy causal mechanisms while retaining sample-count-based sharing for the common diagnostic representation.

\subsection{Interval Counterfactual Fault Reasoning}
\label{sec:counterfactual_reasoning}

After federated training, the learned structural equations allow SeaCausal-FL to estimate how the predicted fault risk changes under a feasible physical intervention. The causal interpretation assumes that the physics-constrained candidate graph provides an approximately valid causal ordering for the retained physical variables, that unmeasured confounding does not dominate the modeled relations, and that the structural mechanisms remain stable within a counterfactual query. Accordingly, causal effects estimated from the real observational data are interpreted as model-based intervention analyses rather than observed causal ground truth; quantitative causal accuracy is evaluated on the real-data-calibrated semi-synthetic benchmark, where the underlying structure and intervention outcomes are known. The counterfactual procedure follows the abduction--action--prediction paradigm while retaining the factual temporal representation and operating-mechanism assignment.

\subsubsection{Abduction of Exogenous Residuals}
\label{sec:counterfactual_abduction}

A counterfactual prediction should describe the same factual sample under a different action. Let
$\mathcal{F}_{j}^{(r)}(\mathbf{s})$
denote the deterministic structural function of node $j$ under mechanism $r$, corresponding to the right-hand side of \eqref{eq:mechanism_structural_equation} without the exogenous residual. For each mechanism, SeaCausal-FL infers the sample-specific residual as
$\mathbf{u}_{k,n}^{(r)}
=
\mathbf{s}_{k,n}
-
\widehat{\mathbf{s}}_{k,n}^{(r)}$.
Equivalently, for an endogenous node $j$,

\begin{equation}
u_{k,n,j}^{(r)}
=
s_{k,n,j}
-
\mathcal{F}_{j}^{(r)}
\left(
\mathbf{s}_{k,n}
\right).
\label{eq:counterfactual_abduction}
\end{equation}

The residual $\mathbf{u}_{k,n}^{(r)}$ represents the sample-specific variation that is not explained by the deterministic structural equations. It is retained during counterfactual propagation so that the factual and counterfactual states correspond to the same underlying sample.

\subsubsection{Action and Causal Propagation}
\label{sec:counterfactual_action}

Consider the intervention $\operatorname{do}(s_{k,n,a}=v)$, where $a$ belongs to the actionable set comprising engine speed, water-brake load, charge-air intercooler cooling-water flow, and engine cooling-water flow. These variables correspond to controllable operating or cooling conditions in the considered system. Their feasible intervention values are obtained from the training-only empirical quantiles $\{0.1,0.3,0.5,0.7,0.9\}$ rather than manually specified ranges. The mechanism assignment is anchored to the factual operating context throughout a counterfactual query. Therefore, interventions on context-defining variables are interpreted as within-mechanism perturbations rather than transitions to a different operating mechanism.

\begin{equation}
s_{k,n,j}^{\mathrm{cf},(r)}
=
\begin{cases}
v,
&
j=a,
\\[1mm]
s_{k,n,j},
&
j\notin\operatorname{De}(a)\cup\{a\},
\\[1mm]
\mathcal{F}_{j}^{(r)}
\left(
\mathbf{s}_{k,n}^{\mathrm{cf},(r)}
\right)
+
u_{k,n,j}^{(r)},
&
j\in\operatorname{De}(a).
\end{cases}
\label{eq:counterfactual_propagation}
\end{equation}

Variables that are neither intervened upon nor causally downstream of the intervention retain their factual values. Equation~\eqref{eq:counterfactual_propagation} produces one counterfactual physical state $\mathbf{s}_{k,n}^{\mathrm{cf},(r)}$ for each mechanism.

The mechanism-specific counterfactual causal logits are obtained by replacing the factual state in \eqref{eq:mechanism_specific_causal_logits} with $\mathbf{s}_{k,n}^{\mathrm{cf},(r)}$, giving

\begin{equation}
\mathbf{z}_{k,n}^{\mathrm{cf},(r)}
=
\mathbf{q}
\left(
\mathbf{s}_{k,n}^{\mathrm{cf},(r)}
\right)
+
\mathbf{W}_{r}^{\mathrm{res}}
\mathbf{s}_{k,n}^{\mathrm{cf},(r)}
+
\mathbf{b}_{r}^{\mathrm{res}}.
\label{eq:mechanism_counterfactual_logits}
\end{equation}

The shared temporal logits $\mathbf{z}_{k,n}^{\mathrm{base}}$ remain factual because the intervention is applied only to the physical causal state rather than to a synthetically generated sensor window.

\subsubsection{Interval Risk Estimation}
\label{sec:counterfactual_interval}

The mechanism assignment is determined from the factual operating context and is not recomputed after intervention. This prevents an intervention from artificially reassigning the sample to another operating mechanism.

Let
$\underline{\boldsymbol{\pi}}_{k,n}$,
$\boldsymbol{\pi}_{k,n}$, and
$\overline{\boldsymbol{\pi}}_{k,n}$
denote the factual lower, midpoint, and upper mechanism-weight vectors. The feasible IT2 weight set is

\begin{equation}
\mathcal{W}_{k,n}
=
\left\{
\mathbf{w}\in\mathbb{R}_{+}^{R}
\;\middle|\;
\begin{aligned}
\mathbf{1}^{\mathsf{T}}\mathbf{w}
&=1,
\\
\underline{\boldsymbol{\pi}}_{k,n}
&\preceq
\mathbf{w}
\preceq
\overline{\boldsymbol{\pi}}_{k,n}
\end{aligned}
\right\}.
\label{eq:counterfactual_weight_set}
\end{equation}
where $\preceq$ denotes componentwise inequality.

For each mechanism, the counterfactual causal logits are first fused with the factual temporal logits and mapped to a class-probability vector. Specifically,
\begin{equation}
\mathbf{p}_{k,n}^{\mathrm{cf},(r)}
=
\mathcal{S}
\left(
\mathbf{z}_{k,n}^{\mathrm{base}}
+
\lambda_{\mathrm{c}}
\mathbf{z}_{k,n}^{\mathrm{cf},(r)}
\right),
\label{eq:mechanism_counterfactual_probability}
\end{equation}
where $\mathcal{S}(\cdot)$ denotes the softmax mapping.
For a feasible mechanism-weight vector
$\mathbf{w}\in\mathcal{W}_{k,n}$, the counterfactual
probability vector is
\begin{equation}
\mathbf{p}_{k,n}^{\mathrm{cf}}(\mathbf{w})
=
\sum_{r=1}^{R}
w_r
\mathbf{p}_{k,n}^{\mathrm{cf},(r)}.
\label{eq:counterfactual_probability_mixture}
\end{equation}
Let $p_{k,n,c}^{\mathrm{cf}}(\mathbf{w})$ denote the $c$th element of $\mathbf{p}_{k,n}^{\mathrm{cf}}(\mathbf{w})$. The nominal and interval-valued counterfactual risks are

\begin{equation}
\begin{aligned}
p_{k,n,c}^{\mathrm{cf,mid}}
&=
p_{k,n,c}^{\mathrm{cf}}
\left(
\boldsymbol{\pi}_{k,n}
\right),
\\
\underline{p}_{k,n,c}^{\mathrm{cf}}
&=
\min_{\mathbf{w}\in\mathcal{W}_{k,n}}
p_{k,n,c}^{\mathrm{cf}}(\mathbf{w}),
\\
\overline{p}_{k,n,c}^{\mathrm{cf}}
&=
\max_{\mathbf{w}\in\mathcal{W}_{k,n}}
p_{k,n,c}^{\mathrm{cf}}(\mathbf{w}).
\end{aligned}
\label{eq:interval_counterfactual_risk}
\end{equation}

Here, $p_{k,n,c}^{\mathrm{cf,mid}}$ is the nominal counterfactual risk, while
$[\underline{p}_{k,n,c}^{\mathrm{cf}},
\overline{p}_{k,n,c}^{\mathrm{cf}}]$
characterizes the variation induced by uncertainty in the factual operating-mechanism assignment. It is therefore interpreted as a mechanism-uncertainty interval rather than a nominal-coverage prediction interval unless an additional calibration procedure is applied.

\section{Experiments}
\subsection{Experimental Setup}
\label{sec:experimental_setup}

\subsubsection{Datasets and Federated Protocol}
We evaluate SeaCausal-FL on the Marine Engine Fault Dataset \cite{bahootoroody2026marine}. Its 15 fault--load recordings are treated as 15 virtual maritime IoT clients and form a six-class task containing normal operation and five fault types. Each recording is chronologically divided into 60\%, 20\%, and 20\% training, validation, and test segments, with a 32-sample guard interval between adjacent segments. Windows of length $T=32$ are extracted with a stride of 8, yielding 6,494/2,018/2,129 windows under the natural partition. Missing values are mean-imputed and accompanied by a binary mask.

Besides the natural partition, we construct an IID partition and Dirichlet label-skew partitions with $\alpha\in\{0.5,1.0\}$. For causal and counterfactual evaluation, we further construct a real-data-calibrated semi-synthetic benchmark from the training portion of the marine-engine dataset. The benchmark retains the 15-client structure and 18 selected physical variables, while the empirical operating ranges, marginal statistics, and fault-related response patterns of the real training data are used to calibrate the synthetic generation process. Mechanism-specific structural causal models generate the physical states and provide known causal coefficient matrices, which are retained as ground truth for causal-structure evaluation. This benchmark is formulated as a binary fault-risk task, and the generated samples are divided into training, validation, and test subsets at the client level. For each factual sample, interventions are applied to predefined actionable variables over feasible values derived from the calibrated operating ranges, and the same generating structural causal model (SCM) is used to obtain the corresponding counterfactual states and fault risks. The generated data are additionally checked against the real-data calibration statistics before causal evaluation. The real dataset is therefore used for the main six-class diagnostic experiments, whereas the real-data-calibrated semi-synthetic benchmark is used only when causal or counterfactual ground truth is required.

\subsubsection{Compared Methods}
For diagnostic evaluation, SeaCausal-FL is compared with six representative federated baselines: a federated TSK fuzzy baseline (FedFuzzy-TSK) implemented following the federated TSK learning setting in \cite{corcuera2023federatedtsk}, FedProx \cite{li2020fedprox}, FedAdam \cite{reddi2021adaptive}, Ditto \cite{li2021ditto}, FedBN \cite{li2021fedbn}, and MOON \cite{li2021moon}. These methods represent fuzzy federated learning, proximal optimization, adaptive server optimization, personalized federated learning, local normalization for feature-shifted non-IID data, and representation-level contrastive learning, respectively. The same temporal backbone and data partitions are used whenever applicable to provide a consistent comparison.

For the causal and counterfactual evaluation on the semi-synthetic benchmark, SeaCausal-FL is further compared with Pooled-SEM, Local-SEM, and Oracle-SEM, which are constructed following standard structural causal modeling principles \cite{peters2017elements}. Pooled-SEM learns a single structural equation model from pooled training samples, Local-SEM estimates separate causal models from individual clients, and Oracle-SEM fits mechanism-specific structural models using the ground-truth operating-mechanism assignments. These baselines respectively represent centralized causal estimation, fully local causal learning, and causal modeling with oracle regime information.

\subsubsection{Implementation Details}
The common backbone concatenates the 69 sensor channels with their missing-value masks, projects the resulting input to 96 channels, and applies two residual one-dimensional convolutional layers with a kernel size of 3, eight-group normalization, GELU activation, and dropout. The average and last-step temporal summaries are combined into a 96-dimensional representation. The diagnostic head contains a 96-unit hidden layer and produces six class logits. FedBN replaces group normalization with client-specific batch normalization. SeaCausal-FL additionally uses three operating-context variables, seven automatically selected fuzzy mechanisms, and an 18-node SCM. Its causal outcome branch contains a shared hidden layer with 96 units and six output logits, together with a mechanism-specific linear residual that maps the 18 physical variables to the six fault classes.

Federated training uses at most 80 rounds, 60\% client participation, two local epochs, and a batch size of 128. AdamW is used with learning rate $8\times10^{-4}$, weight decay $10^{-4}$, gradient clipping at 5, and dropout 0.2. The checkpoint with the highest validation F1-score is used for testing. All fuzzy and causal priors and the intervention quantiles $\{0.1,0.3,0.5,0.7,0.9\}$ are estimated from training data only.

\subsubsection{Evaluation Metrics}
Macro-F1 score is used as the primary diagnostic metric, together with accuracy, precision, recall, AUROC, and AUPRC to evaluate classification and ranking performance. The leave-one-load-out experiment further reports accuracy, precision, recall, and Macro-F1 score to assess generalization to unseen operating conditions. The leave-one-fault-type-out experiment reports the same metrics on the remaining known fault classes after one fault type is completely excluded from training, thereby evaluating diagnostic robustness to incomplete fault-type coverage in the training data. On the real-data-calibrated semi-synthetic causal benchmark, causal structure recovery is evaluated using Edge-F1, Edge-AUPRC, and coefficient RMSE. Counterfactual fidelity is assessed by counterfactual risk MAE, PEHE, and effect-sign accuracy, while policy regret and action accuracy are used to evaluate intervention quality.

\subsection{Performance Evaluation of SeaCausal-FL Framework}
\begin{table*}[t]
\centering
\caption{Diagnostic performance comparison of different models under four client partitions.}
\label{tab:main_diagnostic_comparison}
\setlength{\tabcolsep}{3.0pt}
\renewcommand{\arraystretch}{1.12}
\scriptsize
\begin{tabular}{lcccccccccc}
\toprule
& \multicolumn{4}{c}{Partition-wise F1-score$\uparrow$} & \multicolumn{6}{c}{Average over four partitions} \\
\cmidrule(lr){2-5}\cmidrule(lr){6-11}
Method & IID & Natural Non-IID& Dir. ($\alpha=0.5$) Non-IID& Dir. ($\alpha=1.0$) Non-IID& Acc.$\uparrow$ & Prec.$\uparrow$ & Rec.$\uparrow$ & F1-score$\uparrow$ & AUROC$\uparrow$ & AUPRC$\uparrow$ \\
\midrule
FedFuzzy-TSK & \underline{82.66} & 88.95 & 79.40 & 79.73 & 81.72 & 83.08 & \underline{86.99} & 82.69 & 97.33 & 90.74 \\
FedProx & 77.36 & \underline{89.94} & 86.82 & 77.74 & 83.03 & \underline{83.33} & 86.75 & 83.22 & \underline{98.23} & \underline{91.34} \\
FedAdam & 80.44 & 87.03 & \textbf{88.82} & 77.30 & \underline{84.18} & 83.11 & 85.27 & \underline{83.40} & 97.74 & 88.57 \\
Ditto & 81.70 & 86.65 & 76.12 & \underline{80.94} & 80.77 & 81.41 & 86.53 & 81.35 & 97.31 & 88.29 \\
FedBN & 79.25 & 85.16 & 79.59 & 79.67 & 80.01 & 80.71 & 84.61 & 80.92 & 97.11 & 87.39 \\
MOON & 75.15 & 83.31 & 86.89 & 76.00 & 79.61 & 81.10 & 84.32 & 80.34 & 96.38 & 86.91 \\
\midrule
\textbf{SeaCausal-FL} & \textbf{84.42} & \textbf{90.48} & \underline{87.76} & \textbf{85.61} & \textbf{86.51} & \textbf{86.84} & \textbf{91.23} & \textbf{87.07} & \textbf{98.98} & \textbf{94.81} \\
\bottomrule
\end{tabular}
\vspace{0.6mm}
\end{table*}

Table~\ref{tab:main_diagnostic_comparison} reports the diagnostic results under four client partitions with different degrees of data heterogeneity. SeaCausal-FL achieves the highest F1-score under IID, natural non-IID, and Dirichlet $\alpha=1.0$, reaching 84.42\%, 90.48\%, and 85.61\%, respectively. Under Dirichlet $\alpha=0.5$, FedAdam obtains the highest F1-score of 88.82\%, while SeaCausal-FL achieves 87.76\%, with a difference of only 1.06 percentage points. When the four partitions are considered together, SeaCausal-FL reaches average accuracy, precision, recall, and F1 values of 86.51\%, 86.84\%, 91.23\%, and 87.07\%, respectively. Compared with the strongest baseline for each corresponding metric, the improvements are 2.33, 3.51, 4.24, and 3.67 percentage points. Similar gains are observed for AUROC and AUPRC, where SeaCausal-FL achieves 98.98\% and 94.81\%. In particular, the improvement in recall indicates that fewer fault samples are missed, which is important for marine-engine monitoring where an undetected abnormal condition can lead to a more serious operational consequence.

\begin{figure}[!t]
\centering
\includegraphics[width=\columnwidth]{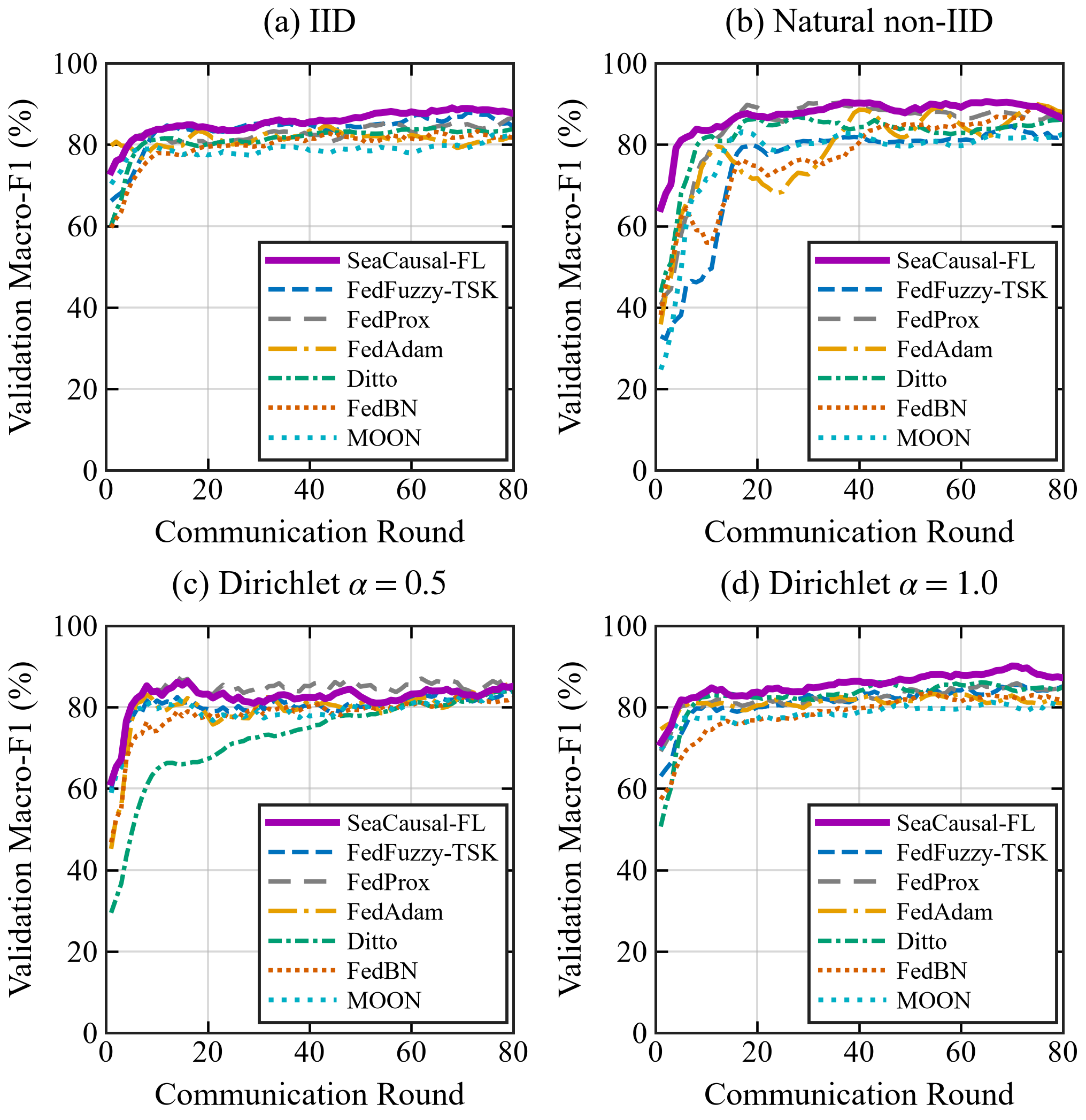}
\caption{Comparison of convergence performance across different models under four client partitions.}
\label{fig:training_curves}
\end{figure}

The convergence behavior in Fig.~\ref{fig:training_curves} provides a further view of the effect of client heterogeneity. Under IID, most methods reach a relatively stable region after the early communication rounds, and their performance differences are moderate. The gap becomes more evident under the natural non-IID setting, where several baselines exhibit larger fluctuations during the first part of training. SeaCausal-FL rises rapidly and remains in a high F1 region after convergence. Similar behavior is observed under the two Dirichlet partitions. In the $\alpha=0.5$ case, several methods remain competitive and FedAdam eventually obtains a slightly higher F1-score, which is consistent with the results in Table~\ref{tab:main_diagnostic_comparison}. However, SeaCausal-FL maintains a relatively stable trajectory as the partition changes. This is particularly relevant to federated marine-engine diagnosis because the composition of participating clients and their local class distributions can vary substantially across communication rounds.

\begin{figure}[!]
\centering
\includegraphics[width=\columnwidth]{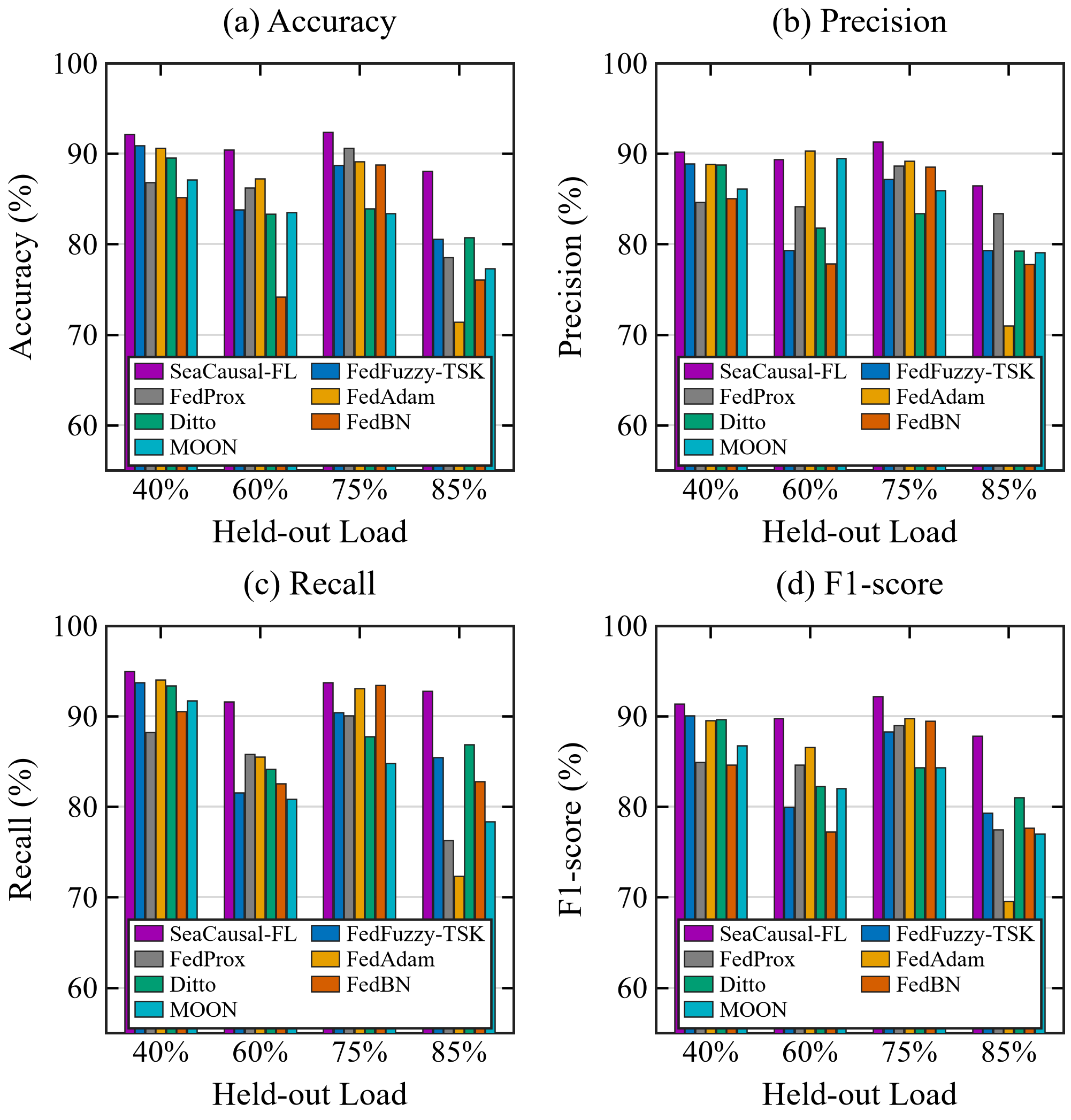}
\caption{Diagnostic performance under leave-one-load-out training.}
\label{fig:unseen_load_metrics}
\end{figure}

The leave-one-load-out experiment further examines whether the learned model remains effective when the operating condition itself is unseen during training. For each setting in Fig.~\ref{fig:unseen_load_metrics}, all samples belonging to one load level are removed from the training set and used only for evaluation. This setting differs from ordinary client heterogeneity because the model must transfer its diagnostic knowledge to an operating region that has not contributed to model optimization. SeaCausal-FL maintains strong accuracy, precision, recall, and F1-score across the four held-out loads, whereas the relative performance of the competing methods changes more noticeably with the excluded load. The difference becomes particularly clear when the 85\% load is held out, where several baselines show a substantial decrease across multiple metrics while SeaCausal-FL retains a comparatively high diagnostic performance.

The advantage under unseen loads is closely related to the treatment of operating conditions in the proposed framework. A diagnostic model trained only from statistical associations can associate a fault with sensor patterns that are specific to the loads represented in the training clients. Once the operating point changes, the same fault can produce a different physical response and the learned decision boundary becomes less reliable. SeaCausal-FL instead represents the operating context through overlapping fuzzy mechanisms and associates each mechanism with its own causal response model. A sample located between two operating regions can therefore receive support from both mechanisms rather than being assigned to a fixed load interval. This soft operating representation reduces abrupt changes between neighboring conditions and provides a smoother basis for transferring the learned physical relations to an unseen load.

\begin{figure}[!t]
\centering
\includegraphics[width=\columnwidth]{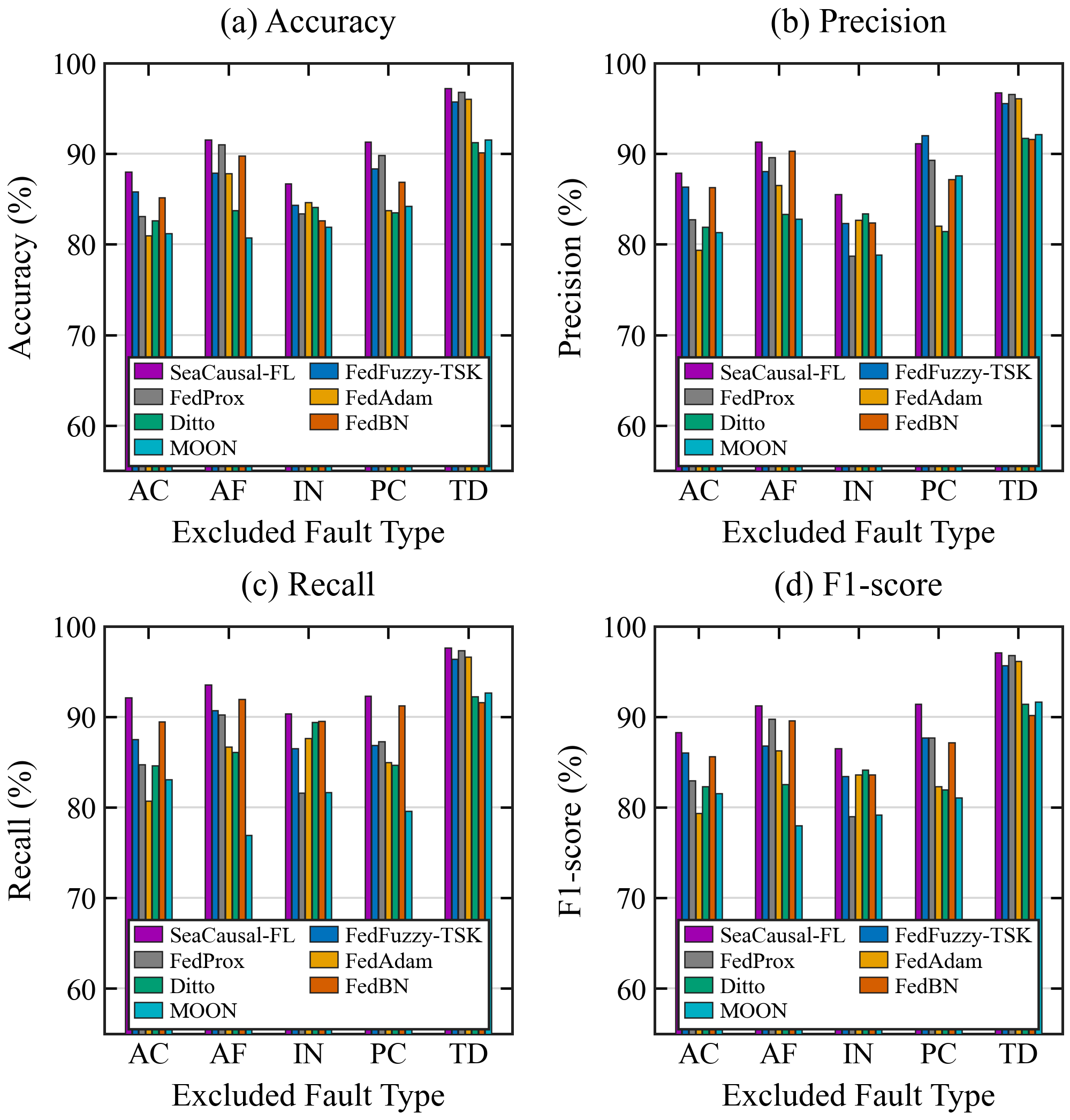}
\caption{Diagnostic performance under leave-one-fault-type-out training.}
\label{fig:leave_one_fault_source_metrics}
\vspace{-1.0em}
\end{figure}
A different form of training-data heterogeneity is examined through the leave-one-fault-type-out evaluation in Fig.~\ref{fig:leave_one_fault_source_metrics}. In each setting, all samples belonging to one fault type are excluded from model training, and diagnostic performance is evaluated on the remaining known fault classes. This experiment therefore examines the sensitivity of federated diagnosis to incomplete fault-type coverage rather than recognition of the excluded fault itself. The performance of several baselines varies noticeably as different fault types are removed, indicating that the composition of fault categories available during training can substantially affect the learned representation.

SeaCausal-FL remains consistently competitive across all five fault-omission settings and shows smaller performance degradation in the more challenging cases. The advantage is observed across accuracy, precision, recall, and F1-score rather than being confined to a single metric, suggesting that the proposed representation is less sensitive to the absence of a particular fault category during training.

This behavior can be attributed to the complementary roles of the two diagnostic paths. The shared temporal encoder extracts fault-discriminative information that can be reused across clients, while the fuzzy causal branch describes how changes in operating state are related to downstream physical responses. The final prediction therefore does not rely solely on a fault pattern observed under a particular client or load condition. Such a representation is less sensitive to changes in the fault-category composition of the training data and helps preserve diagnostic performance when some fault types are unavailable during model optimization.

\subsection{Ablation and Sensitivity Analyses}
\begin{table}[!t]
\centering
\caption{Ablation results under the natural non-IID partition.}
\label{tab:ablation_study}
\setlength{\tabcolsep}{3.0pt}
\renewcommand{\arraystretch}{1.10}
\scriptsize
\resizebox{\columnwidth}{!}{%
\begin{tabular}{lcccccc}
\toprule
Variant
& Acc.$\uparrow$
& Prec.$\uparrow$
& Rec.$\uparrow$
& F1-score$\uparrow$
& AUROC$\uparrow$
& AUPRC$\uparrow$ \\
\midrule

w/o IT2
& 83.33 & 82.90 & 87.47 & 82.62 & 98.11 & 87.50 \\

Single mechanism ($R=1$)
& 88.40
& 86.98
& 90.13
& 88.03
& 98.82
& 91.64 \\

w/o SEM
& 84.78 & 84.33 & 88.95 & 84.36 & 98.19 & 87.70 \\

w/o edge-prior regularization
& 84.78 & 84.41 & 89.17 & 84.39 & 98.28 & 87.91 \\

w/o causal residual
& 84.08 & 83.51 & 88.24 & 83.48 & 98.16 & 87.60 \\

w/o alignment
& 86.38 & 85.69 & 89.20 & 86.44 & 98.04 & 91.68 \\
\midrule

\textbf{SeaCausal-FL}
& \textbf{90.28}
& \textbf{89.84}
& \textbf{93.39}
& \textbf{90.48}
& \textbf{99.02}
& \textbf{92.48} \\
\bottomrule
\end{tabular}%
}
\vspace{-1mm}
\end{table}

Table~\ref{tab:ablation_study} evaluates the contribution of the main components under the natural non-IID partition. Removing the IT2 fuzzy modeling reduces F1-score from 90.48\% to 82.62\% and AUPRC from 92.48\% to 87.50\%. This confirms that the fuzzy layer is not used only for uncertainty estimation; it also represents overlapping operating conditions through soft mechanism memberships rather than fixed regime boundaries \cite{liang2000interval,qiao2024interval}. Removing the SEM loss or edge-prior regularization produces similar reductions, with F1-score decreasing to 84.36\% and 84.39\%, respectively. The former weakens the structural modeling of physical relations, while the latter weakens the constraint that keeps learned edge probabilities close to their data-driven structural priors. Their comparable degradation indicates that both structural reconstruction and selective causal relations are important for the causal branch \cite{pearl2009causal,scholkopf2021toward}.

The context-conditioned causal residual also has a pronounced effect. Without this component, F1-score decreases to 83.48\%, indicating that operating-dependent causal information needs to directly refine the shared diagnostic output. In comparison, using a single causal mechanism gives a smaller reduction to 88.03\%. This result suggests that one common mechanism already captures part of the shared physical structure, but multiple mechanisms remain useful for describing operating-dependent responses. Without mechanism alignment, F1-score decreases to 86.44\%. Local mechanisms can drift toward different operating regions during training, and direct aggregation can consequently combine components with different physical meanings. The result supports mechanism-level coordination before federated aggregation, consistent with the need to coordinate causal knowledge learned from distributed data \cite{yang2024fedcausal,guo2024fedcsl}.

It is also noteworthy that AUROC remains above 98\% for all ablation variants, whereas F1-score and AUPRC change more substantially. This indicates that the shared temporal path preserves strong overall class separability, while the fuzzy causal components mainly improve the reliability of the final decisions under heterogeneous operating conditions.
\begin{figure}[!t]
\centering
\includegraphics[width=\columnwidth]{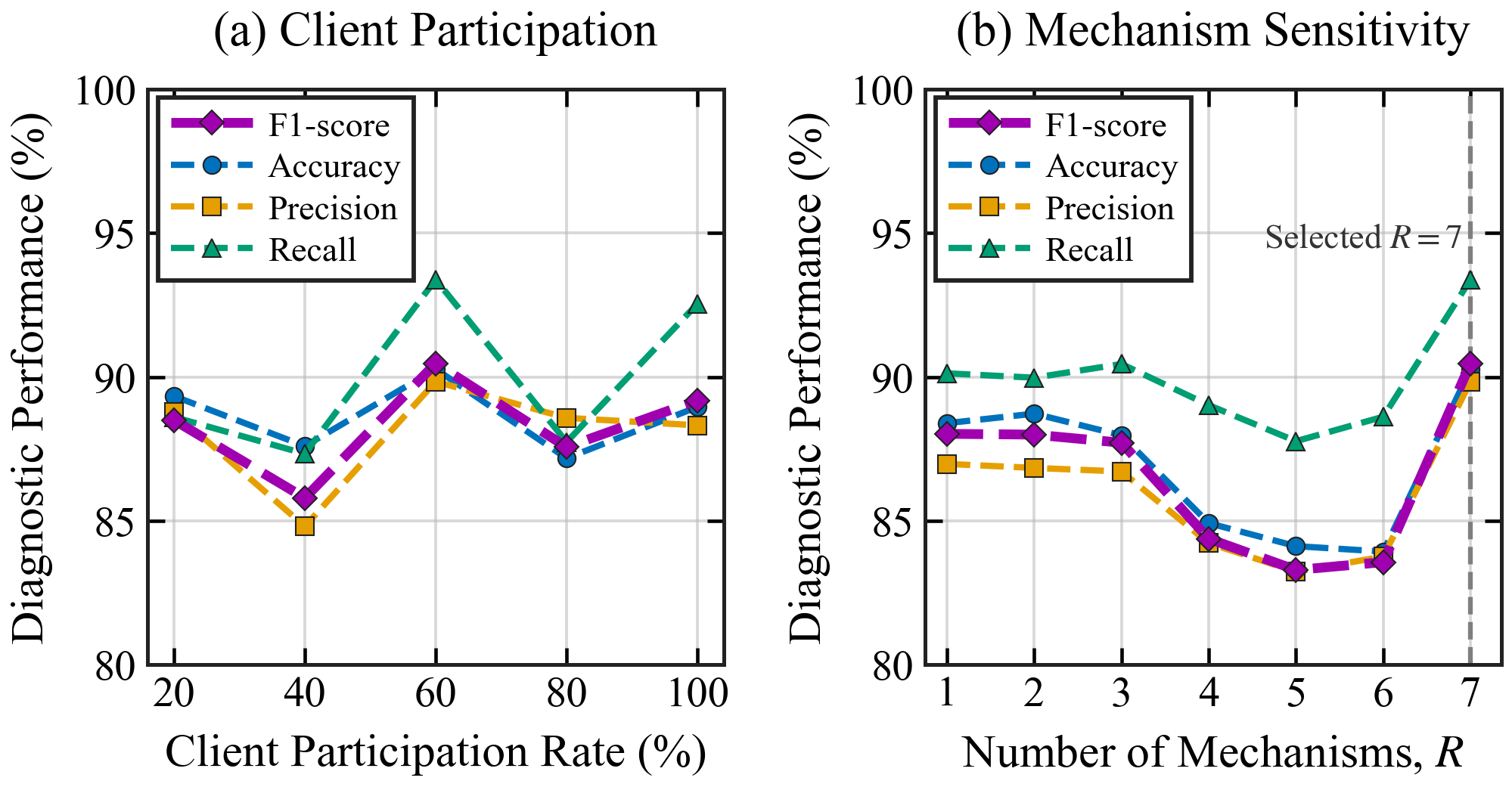}
\caption{Performance under different client participation ratios and causal-mechanism numbers.}
\label{fig:participation_mechanism_sensitivity}
\end{figure}

The effect of client availability is examined in Fig.~\ref{fig:participation_mechanism_sensitivity}(a). F1-score remains approximately between 85\% and 90\% as the participation ratio varies from 20\% to 100\%, with the highest value observed around 60\%. The nonmonotonic trend reflects the tradeoff between information coverage and heterogeneous local updates: fewer participating clients increase round-to-round variation, whereas more clients introduce a broader range of local distributions. More importantly, the overall variation remains moderate, showing that SeaCausal-FL does not rely on full client participation in every communication round.

The number of fuzzy causal mechanisms shows a stronger influence in Fig.~\ref{fig:participation_mechanism_sensitivity}(b). With $R=1$--$3$, F1-score remains close to 88\%, while $R=4$--$6$ produces a noticeable decrease. Dividing the operating space into more mechanisms can reduce the effective samples available for mechanism-specific structural estimation when the resulting decomposition is not sufficiently informative. The sensitivity analysis evaluates the complete set of mechanism numbers satisfying the sample-sufficiency criterion rather than manually truncating the search at $R=7$. For the present training data, $R=1,\ldots,7$ remain eligible, while larger values are excluded because at least one fitted component does not provide sufficient soft or hard sample support for reliable structural estimation. Within this admissible set, $R=7$ is selected by the data-driven criterion and also provides the strongest diagnostic performance. This agreement supports determining the mechanism number from the available operating evidence rather than prescribing a fixed number of regimes.

\subsection{Causal Interpretation and Counterfactual Reasoning}
\label{sec:causal_counterfactual_results}

\begin{figure}[!t]
\centering
\includegraphics[width=\columnwidth]{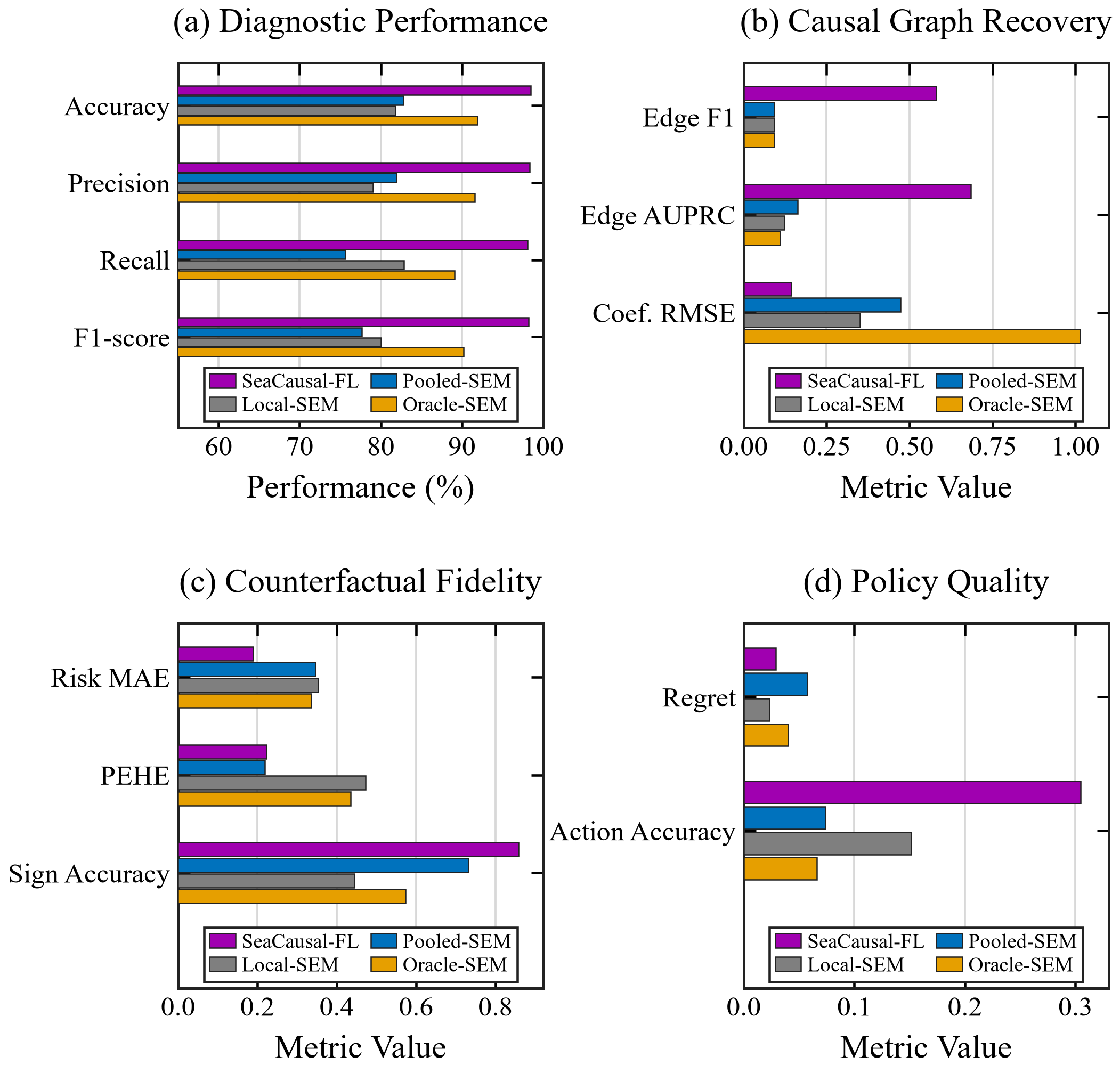}
\caption{Comparison of diagnostic performance, causal graph recovery, counterfactual fidelity, and policy quality across different methods.}
\label{fig:causal_counterfactual_benchmarks}
\end{figure}

The causal evaluation is conducted on the real-data-calibrated semi-synthetic benchmark, where the ground-truth causal structure and intervention outcomes are available for evaluation. This setting allows diagnostic performance, causal structure recovery, counterfactual estimation, and intervention quality to be examined within the same controlled environment. As shown in Fig.~\ref{fig:causal_counterfactual_benchmarks}(a), SeaCausal-FL achieves the strongest factual diagnostic performance among the compared causal models in terms of accuracy, precision, recall, and F1-score. This result indicates that introducing mechanism-specific causal modeling does not compromise the shared diagnostic capability. Instead, the causal branch provides operating-dependent corrections while the shared temporal path preserves fault-discriminative information.

\begin{figure}[!t]
\centering
\includegraphics[width=\columnwidth]{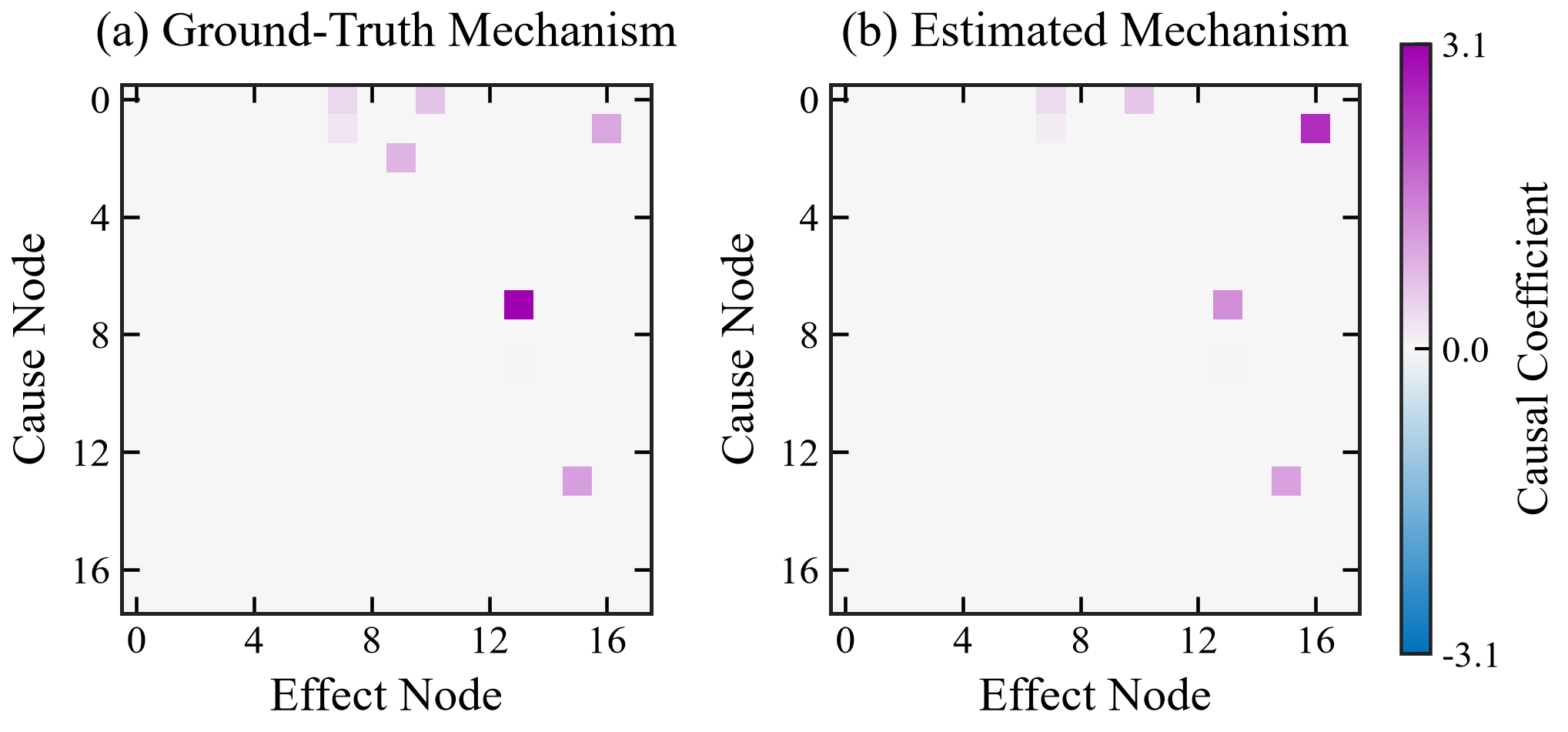}
\caption{Comparison between the ground-truth and estimated causal coefficient matrices for a representative mechanism.}
\label{fig:causal_mechanism_comparison}
\end{figure}

The learned causal structure is further examined qualitatively in Fig.~\ref{fig:causal_mechanism_comparison} and quantitatively in Fig.~\ref{fig:causal_counterfactual_benchmarks}(b). For the representative operating mechanism, the estimated coefficient matrix recovers the main sparse pattern and several dominant relations in the ground-truth matrix, although differences remain in individual coefficient magnitudes. SeaCausal-FL achieves an Edge-F1 of approximately 0.58 and an Edge-AUPRC of 0.68, while the compared SEM baselines remain below approximately 0.10 and 0.17, respectively. Its coefficient RMSE is also reduced to about 0.14. The higher edge-level accuracy and lower coefficient error indicate that the proposed model recovers a more selective and quantitatively accurate structural representation. This is important for subsequent intervention analysis because counterfactual propagation depends on both directed relations and their effect magnitudes \cite{pearl2009causal,scholkopf2021toward}. The mechanism-specific formulation also differs from federated causal approaches that mainly recover a single global causal structure from distributed data \cite{yang2024fedcausal,guo2024fedcsl}.

The counterfactual results in Fig.~\ref{fig:causal_counterfactual_benchmarks}(c) examine whether the recovered structural relations can reproduce intervention effects at the sample level. SeaCausal-FL obtains the lowest counterfactual risk MAE among the compared methods and remains competitive in PEHE. It also achieves the highest effect-sign accuracy, indicating that the predicted direction of risk change agrees more frequently with the reference intervention effect. Correctly identifying whether an intervention increases or decreases fault risk is particularly relevant to maintenance reasoning. The larger PEHE of Local-SEM and Oracle-SEM further shows that fitting separate structural models to local data or known operating regimes alone does not necessarily provide accurate individual-level counterfactual effects.

The policy-level results in Fig.~\ref{fig:causal_counterfactual_benchmarks}(d) further evaluate whether these counterfactual estimates support useful intervention selection. SeaCausal-FL achieves the highest action accuracy, indicating that its selected intervention agrees with the reference optimal action more frequently than those of the SEM baselines. Its policy regret is also lower than those of Pooled-SEM and Oracle-SEM and remains close to the lowest value obtained by Local-SEM. The difference between these two metrics is expected: action accuracy evaluates whether the exact optimal action is selected, whereas policy regret measures the excess cost associated with the selected action. A method can therefore obtain relatively low regret by selecting a near-optimal action even when it does not exactly match the reference optimum. Overall, SeaCausal-FL maintains a favorable balance between causal structure recovery, sample-level counterfactual estimation, and intervention selection.

\section{Conclusion}
This paper presented SeaCausal-FL for federated marine-engine fault diagnosis under operating-condition and label heterogeneity. The framework combines a shared temporal diagnostic model with data-driven interval type-2 fuzzy mechanisms and physics-constrained structural causal models. Mechanism alignment and evidence-aware aggregation preserve operating-specific knowledge across clients, while the learned structural equations enable intervention-based counterfactual reasoning. Experiments on the real marine-engine dataset showed strong performance across IID and non-IID partitions, unseen loads, and incomplete fault-type coverage. Ablation results verified the importance of fuzzy mechanism modeling, causal structural learning, residual correction, and mechanism alignment. On the real-data-calibrated causal benchmark, SeaCausal-FL also achieved more accurate causal-structure recovery and favorable counterfactual and policy-level performance. Future work will consider larger multi-vessel deployments and richer real intervention data. As a potential future direction, we are looking forward to extending our method to improve the performance of various applications, such as large language
models~\cite{lin2024splitlora,duan2025llm,fang2026hfedmoe} and split learning
system~\cite{lin2025hierarchical,wei2025optimizing,fang2026nsc,lin2025hasfl}.

\ifCLASSOPTIONcaptionsoff
  \newpage
\fi

{\footnotesize
\bibliographystyle{IEEEtran}
\bibliography{IEEEabrv,Bibliography}}

\vfill

\end{document}